\documentclass[sigconf, nonacm]{acmart}
\usepackage{makecell}
\usepackage{algorithm}
\usepackage{algpseudocode}
\usepackage[most]{tcolorbox}
\usepackage{booktabs}
\usepackage{multirow}
\usepackage{xcolor}
\usepackage{colortbl}
\usepackage{siunitx}
\usepackage{tabularx}
\usepackage{array}
\usepackage{ragged2e} 
\usepackage{balance}  
\usepackage{stfloats} 
\newcommand{\ph}[1]{\textcolor{gray}{\texttt{\{#1\}}}}
\newcommand{\pcode}[1]{{\ttfamily #1}}
\newcommand{\promptcell}[1]{%
  \begingroup%
  \footnotesize%
  \ttfamily%
  \setlength{\parskip}{2pt}%
  #1%
  \endgroup%
}

\newcommand{\eg}{\emph{e.g., }}
\newcommand{\ie}{\emph{i.e., }}

\newcommand{\eat}[1]{}

\newcommand{\best}[1]{\textcolor{black}{\textbf{#1}}}   
\newcommand{\negv}[1]{\textcolor{negval}{#1}}
\definecolor{bestval}{HTML}{1A6B3C}    
\definecolor{negval}{HTML}{C0392B}     
\definecolor{rowalt}{HTML}{F5F7FA}     
\definecolor{oursrow}{HTML}{EBF5FB}    

\newcommand\vldbdoi{XX.XX/XXX.XX}

\newcommand\vldbpagestyle{plain} 

\begin{document}
\title{ProfiLLM: Utility-Aligned Agentic User Profiling for Industrial Ride-Hailing Dispatch}







\author{Tengfei Lyu}
\authornote{Equal contribution. Work done during internship at Didichuxing Co. Ltd.}  
\affiliation{%
  \institution{HKUST(GZ)}
  \country{}}
\email{tlyu077@connect.hkust-gz.edu.cn}

\author{Zirui Yuan}
\authornotemark[1]
\affiliation{%
  \institution{HKUST(GZ)}
  \country{}}
\email{zyuan779@connect.hkust-gz.edu.cn}

\author{Xu Liu}
\affiliation{%
  \institution{Didichuxing Co. Ltd}
  \country{}}
\email{leoliuxu@didiglobal.com}

\author{Kai Wan}
\affiliation{%
  \institution{Didichuxing Co. Ltd}
  \country{}}
\email{peterwan@didiglobal.com}

\author{Zihao Lu}
\affiliation{%
  \institution{Didichuxing Co. Ltd}
  \country{}}
\email{luzihao@didiglobal.com}

\author{Li Ma}
\affiliation{%
  \institution{Didichuxing Co. Ltd}
  \country{}}
\email{malimarey@didiglobal.com}

\author{Hao Liu}
\authornote{Corresponding author.}
\affiliation{%
  \institution{HKUST(GZ)}
  \country{}}
\email{liuh@ust.hk}

\begin{abstract}
Bringing Large Language Models (LLMs) into industrial ride-hailing dispatch as semantic feature extractors over platform-scale behavioral logs is a compelling but under-explored data systems problem. Production matching pipelines remain dominated by structured numerical features, yet decisive behavioral signals (e.g., a driver's habitual aversion to certain regions) are inherently contextual and naturally expressible as LLM-generated user profiles.
However, scaling such profiling to a live, millisecond-latency dispatcher faces three intertwined constraints rarely addressed together: on a platform with millions of daily orders, logs exceed any LLM's context window by orders of magnitude; most users are long-tail, with too few interactions for per-user profiling; and surface-fluent profiles do not necessarily improve downstream prediction utility.
%
%
We present ProfiLLM, an agentic LLM data pipeline that operationalizes utility-aligned user profiling for production matching systems through two modules. (1) Tool-Augmented Global Knowledge Mining equips an LLM agent with 27 analytical tools to mine platform-scale data, producing reusable global knowledge, adaptive user clustering rules, and region-level supply-demand priors. (2) Utility-Aligned Profile Exploration generates multiple candidate profiles per cluster, evaluates them via a lightweight downstream utility proxy, iteratively refines the best candidates and constructs preference pairs for DPO fine-tuning.
The pipeline enforces a strict offline--online contract: all LLM reasoning stays offline, while online serving reduces to a lookup of pre-computed cluster-level profile embeddings, adding sub-millisecond overhead with zero online LLM inference.
Deployed on DiDi's production dispatcher, ProfiLLM achieves up to +6.14\% relative AUC improvement in outcome prediction, up to +4.35\% GMV gain in dispatching simulation, and consistent improvements in a 14-day online A/B test including +0.47\% GMV, +0.33\% Completion Rate, and $-$0.82\% Cancel-Before-Accept rate.
Code, extended experiments, additional analyses, and supplementary materials are available at our project page \url{https://ProfiLLM.github.io}.
\end{abstract}


\maketitle

\pagestyle{\vldbpagestyle}




\begin{figure}[!t]
    \centering
    \includegraphics[width=\linewidth]{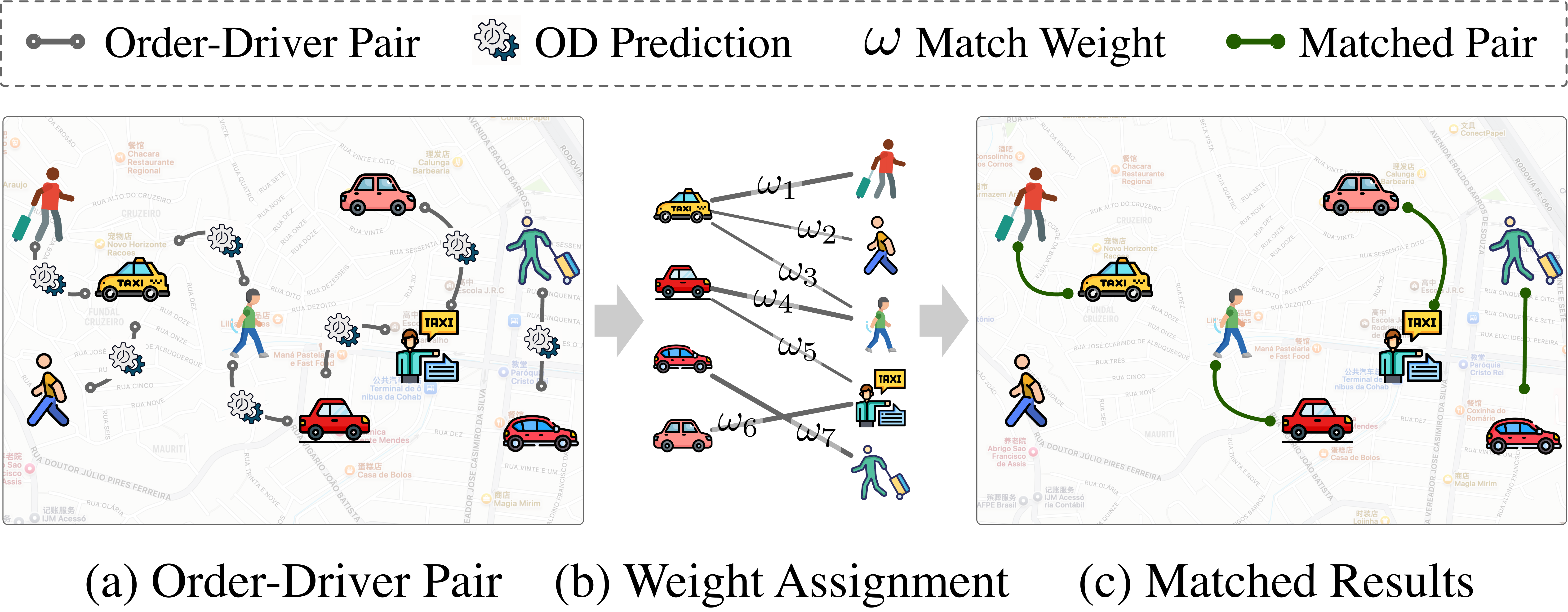}
    \vspace{-0.6cm}
    \caption{Industrial order-dispatching pipeline: (a) outcome prediction for candidate order-driver (OD) pairs, (b) weight assignment, and (c) bipartite matching. Our work improves stage (a) via LLM-generated user profiles.}
    \Description{A three-stage pipeline diagram showing how candidate order-driver pairs flow from outcome prediction, through weight assignment, to bipartite matching. The first stage is highlighted to indicate where LLM-generated user profiles are injected to augment structured features.}
    \label{fig:task}
    \vspace{-0.6cm}
\end{figure}

\section{Introduction}
\label{sec:introduction}

Ride-hailing services have become an essential component of modern urban transportation, fundamentally changing how people commute and travel in cities worldwide.
At the core of these platforms is order dispatching, which continuously matches passenger requests with available drivers under stringent real-time latency constraints.
To evaluate the quality of each potential match, the platform must anticipate a sequence of user behavioral outcomes along the order fulfillment funnel for each candidate order-driver (OD) pair, including whether the driver will \emph{accept} the dispatched order and whether either party will \emph{cancel} after acceptance but before trip completion.
As illustrated in Figure~\ref{fig:task}, a typical production pipeline operates in three stages.
(i) Predicting these per-stage outcomes for each candidate OD pair.
(ii) Composing the predicted probabilities into matching weights that quantify the expected utility of each assignment.
(iii) Computing a globally optimal assignment via bipartite matching, \eg Kuhn-Munkres.
Among these stages, outcome prediction is the primary quality bottleneck, as prediction errors at any stage propagate directly into suboptimal matches that increase passenger waiting time, reduce driver income efficiency, and degrade platform revenue.

Current production outcome predictors mainly rely on structured numerical features (\eg distance, price), leaving implicit semantic factors largely unexploited. 
However, decisive signals governing acceptance and cancellations are inherently contextual, such as a driver's long-term aversion to certain areas or a passenger's strict time-sensitivity during weekday mornings.
Such patterns are difficult to capture via handcrafted numerical features, but they can be naturally expressed as user profiles in language.
In particular, Large Language Models (LLM) possess strong summarization and reasoning capabilities that enable them to distill complex behavioral trajectories into semantically rich contextual profiles, motivating the exploration of LLM-based user profiling for outcome prediction.

\begin{figure}[t]
    \centering
    \includegraphics[width=\linewidth]{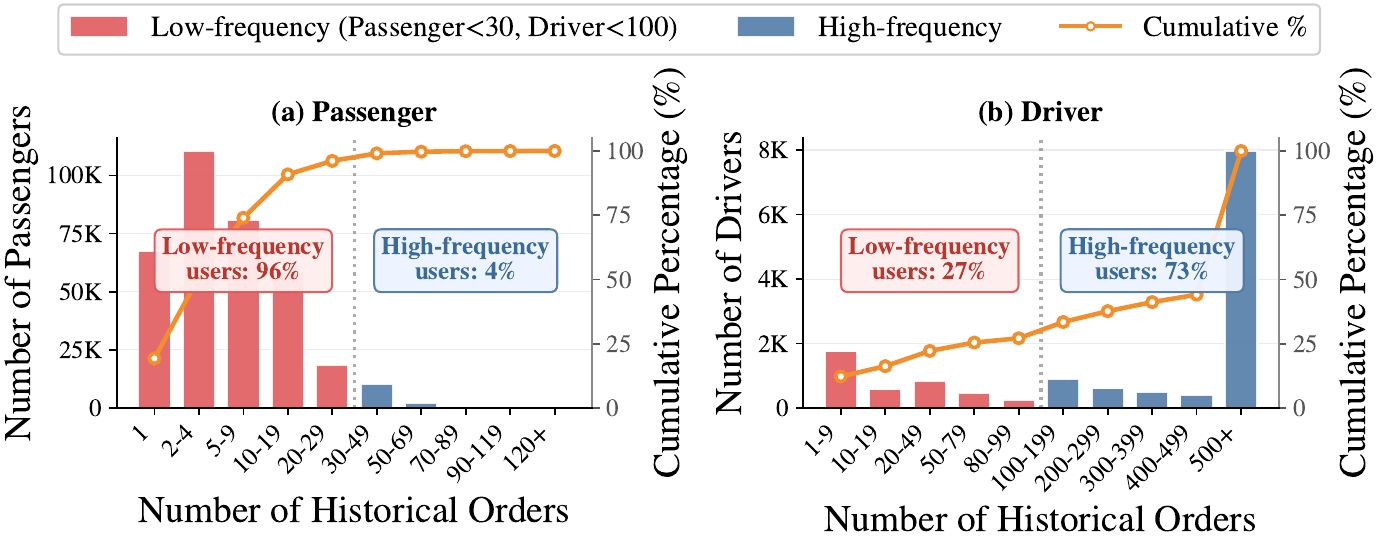}
    \vspace{-0.6cm}
    \caption{Distribution of historical order counts over a 38-day period in City A, revealing a severe long-tail pattern.}
    \Description{A histogram showing the distribution of per-user historical order counts over a 38-day window in City A. The mass is concentrated at small counts and decays rapidly, exhibiting a heavy long-tail with the majority of users contributing only a few orders.}
    \label{fig:longtail}
    \vspace{-0.6cm}
\end{figure}

To validate this hypothesis, we conducted a pilot study on high-frequency users who have sufficient historical interactions. 
We employed an LLM to summarize their historical order trajectories into contextual profiles and incorporated it as additional features for outcome prediction. 
Averaged across high-frequency users, the profile-enhanced approach achieves \textbf{3.71\%} and \textbf{9.64\%} relative AUC improvement for driver cancellation and passenger cancellation, respectively, over models using only structured features, indicating that LLM-generated contextual profiles capture OD-relevant signals largely invisible to traditional feature-based methods.

However, the pilot study was conducted offline and restricted to high-frequency users with rich order histories. While these offline gains are encouraging, bridging the gap to production-scale online deployment introduces several fundamental obstacles.
First, the majority of users are low-frequency and lack sufficient order history for reliable LLM-based profiling. 
Second, without platform-level knowledge to ground the generation, the LLM lacks a consistent reference frame, and when conditioned on a single user's sparse trajectory, it produces profiles of highly variable quality.
Third, even when order history is abundant, the generated profiles are not guaranteed to improve downstream prediction, as the LLM has no explicit signal to optimize for prediction utility. 
These observations motivate three interconnected challenges that must be addressed.

\textbf{Challenge 1: Scalably mining global operational knowledge from massive historical data.}
Generating reliable knowledge requires understanding platform-level regularities such as temporal patterns, regional heterogeneity, and causal factors underlying order outcomes. Such global knowledge provides grounding context that individual user histories alone cannot supply, yet raw logs are massive, a small city in our deployment yields 44.3M dispatching records over a 38-day window, far exceeding any LLM's context capacity, and a full multi-city refresh scales by another order of magnitude.
\textbf{Challenge 2: Adaptive user clustering under long-tail data distributions.}
Even with global knowledge, per-user profiles remain infeasible for the dominant low-frequency population. Over the same 38-day window in City~A, 96\% of passengers appear in $\leq 30$ orders (Figure~\ref{fig:longtail}); while drivers exhibit higher engagement (73\% high-frequency), the passenger-side sparsity fundamentally limits user profiling since outcomes from both parties must be predicted for each candidate match. A principled clustering mechanism is required to group users by behavioral similarity so that each cluster accumulates sufficient data for reliable profile learning.
\textbf{Challenge 3: Ensuring LLM-generated profiles are utility-aligned with downstream prediction.}
With global knowledge and user clusters, LLMs can still produce fluent profile descriptions that fail to capture the specific factors driving behavioral decisions. As our experiments later confirm (Table~\ref{tab:auc_results}), several strong off-the-shelf LLM backbones, degrade downstream prediction AUC by up to $-7.57\%$ when used naively, demonstrating that profile fluency is not a reliable proxy for prediction utility. A systematic mechanism is therefore needed to explore, evaluate, and refine profiles based on measurable downstream utility, while remaining compatible with strict online latency constraints.

To address these challenges, we introduce \textbf{ProfiLLM}, a practical data framework for deploying utility-aligned LLM user profiles in real-time ride-hailing matching. 
%
We devise a \emph{Tool-Augmented Global Knowledge Mining} module in which an LLM agent, equipped with 27 analytical tools, analyzes platform-scale historical data following an \texttt{Explore $\rightarrow$ Deepen $\rightarrow$ Validate $\rightarrow$ Synthesize} paradigm, producing (i) actionable global knowledge (\eg, temporal patterns, weather impacts, causal relationships), (ii) adaptive user clustering with interpretable rules at appropriate granularity, and (iii) regional supply-demand priors from grid-level spatial analyses.
We propose a \emph{Utility-Aligned Profile Exploration} mechanism that, for each user cluster, generates candidates profiles conditioned on the mined global knowledge, scores them via rule-based prediction as a lightweight utility proxy, and iteratively refines the best candidates through feedback derived from prediction error analysis. The resulting utility comparisons yield preference pairs that drive DPO fine-tuning, further improving profile generation quality.
The generated profiles are embedded and integrated with structured features for real-time outcome prediction. 
Critically, all LLM inference occurs offline; the online system operates solely on pre-computed cluster-level profile embeddings, ensuring negligible added latency compatible with production requirements.
Our contributions:
\begin{itemize}

\item We design \textbf{ProfiLLM}, an agentic LLM data pipeline that enforces a strict offline--online contract for behavioral profiling in industrial ride-hailing: all LLM reasoning is batch-offline, while the production dispatcher serves only pre-computed cluster-level profile embeddings. To our knowledge, this is the first deployed LLM-based user profiling pipeline for a production ride-hailing dispatcher.
\item We design a \textbf{Tool-Augmented Global Knowledge Mining} module that exposes 27 analytical tools as a composable operator layer, driven by an LLM agent under an Explore-Deepen-Validate-Synthesize paradigm. The agent autonomously composes operator chains to extract actionable global knowledge, adaptive user-clustering rules, and regional supply-demand priors from platform-scale logs that exceed any LLM context window.
\item We propose a \textbf{Utility-Aligned Profile Exploration} mechanism that turns prediction utility into a first-class optimization signal: for each user cluster, multiple candidate profiles are generated and ranked by a lightweight LOGIC-rule proxy of downstream prediction; the resulting cross-candidate preferences drive DPO fine-tuning, aligning profile generation with the downstream matching objective.
\item We \textbf{deploy ProfiLLM} on DiDi's production dispatcher. The online path adds sub-millisecond overhead with zero online LLM inference. ProfiLLM yields up to \textbf{+6.14\%} outcome-prediction AUC and \textbf{+4.35\%} simulator GMV; a 14-day City~A A/B test confirms \textbf{+0.47\%} GMV, \textbf{+0.33\%} completion rate, and \textbf{$-$0.82\%} cancel rate.

\end{itemize}

\begin{figure*}[t]
    \centering
    \includegraphics[width=\textwidth]{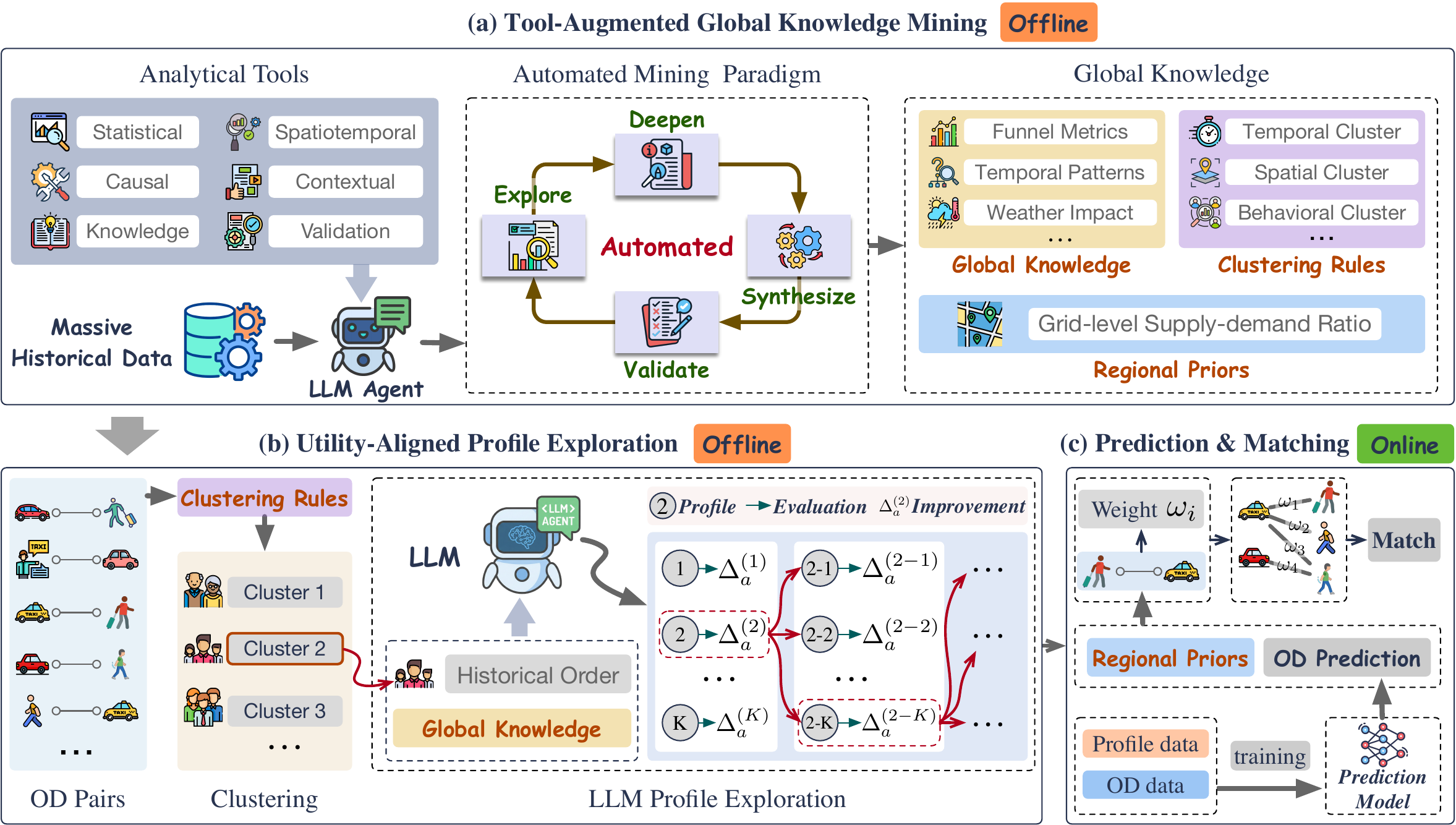}
    \vspace{-0.5cm}
    \caption{Overview of ProfiLLM. (a) Tool-Augmented Global Knowledge Mining and (b) Utility-Aligned Profile Exploration run offline; (c) is the online serving path.}
    \Description{A three-panel architectural diagram of the ProfiLLM framework, with offline badges on stages (a) and (b) and an online badge on stage (c). Panel (a) shows an LLM agent invoking analytical tools over historical data to produce global knowledge, clustering rules, and regional priors. Panel (b) shows an iterative explore-evaluate-refine loop that produces cluster-level user profiles aligned with downstream prediction utility. Panel (c) shows how the generated profiles are encoded as embeddings and fused with structured features for online outcome prediction and bipartite matching.}
    \label{fig:framework}
\end{figure*}

\section{Problem Formulation}
\label{sec:problem}

Consider a ride-hailing platform operating in a city partitioned into a set of spatial grids $\mathcal{G} = \{g_1, g_2, \ldots, g_{|\mathcal{G}|}\}$. Let $\mathcal{P} = \{p_1, p_2, \ldots, p_M\}$ denote the set of passengers and $\mathcal{D} = \{d_1, d_2, \ldots, d_N\}$ denote the set of drivers. Each user $u \in \mathcal{P} \cup \mathcal{D}$ has a historical interaction sequence $\mathcal{H}_u = \{h_1^u, h_2^u, \ldots, h_{|\mathcal{H}_u|}^u\}$, where each record $h_i^u$ contains order-level information such as timestamp, location, and trip details.

At each dispatching cycle, the system receives pending orders $\mathcal{O}$ and identifies candidate matches $\mathcal{C} \subseteq \mathcal{O} \times \mathcal{D}$. 
For each candidate pair $c = (o, d)$ with associated passenger $p$, the system predicts multiple order-level outcomes (\ie order acceptance) that are used to compute matching weights. 
Let $\mathbf{y}_c = (y_c^{(1)}, y_c^{(2)}, \ldots, y_c^{(L)})$ denote the outcome vector, where $L$ is the number of prediction tasks and each $y_c^{(l)} \in \{0, 1\}$ indicates whether the $l$-th stage in the order fulfillment process succeeds (\eg $l=1$ for order acceptance, $l=2$ for order completion). The prediction model estimates: $\hat{\mathbf{y}}_c = f(\mathbf{x}_c, \mathbf{e}_p, \mathbf{e}_d)$, where $\hat{\mathbf{y}}_c = (\hat{y}_c^{(1)}, \ldots, \hat{y}_c^{(L)})$ with $\hat{y}_c^{(l)} \in [0,1]$ being the predicted probability for outcome $l$, $\mathbf{x}_c$ denotes structured features (ETA, distance, price) and $\mathbf{e}_p, \mathbf{e}_d$ are profile embeddings.

These predictions are aggregated into a matching weight $w_c$ quantifying the expected utility of each assignment. The optimal matching $\mathcal{M}^*$ that maximizes $\sum_{c \in \mathcal{M}} w_c$ is then computed via the Kuhn-Munkres algorithm, subject to one-to-one matching constraints. Our work improves outcome prediction by generating utility-aligned user profiles whose embeddings $\mathbf{e}_p$ and $\mathbf{e}_d$ complement structured features.

\section{Methodology}
\label{sec:method}

\subsection{System Overview}
\label{subsec:overview}

\textbf{Production setting.} ProfiLLM is designed and deployed in DiDi's production dispatching service, whose operating constraints shape every design choice below. The dispatcher operates in fixed 2-second cycles; in each cycle, it enumerates candidate order--driver (OD) pairs, predicts per-pair behavioral outcomes, composes them into matching weights, and solves the assignment via the Kuhn--Munkres (KM) algorithm under a $\sim$200\,ms end-to-end latency budget. The production predictor is a multi-task deep model producing calibrated probabilities for four behavioral events along the fulfillment funnel: driver acceptance, post-acceptance driver/passenger cancellation, and completion. Since LLM inference at per-OD-pair scale is incompatible with this budget, ProfiLLM materializes a strict offline--online decoupling as the three-layer pipeline in Figure~\ref{fig:framework}.

\textbf{Layer 1 --- Tool-Augmented Global Knowledge Mining (offline).} An LLM agent equipped with 27 analytical tools mines platform-scale historical logs to produce three reusable artifacts: global behavioral knowledge $\mathcal{K}$, an interpretable user-clustering rule set $\mathcal{A}$, and regional supply--demand priors $\mathcal{R}$. This layer runs as a batch job over the log warehouse; tools operate as composable analytical operators that the agent chains autonomously under an Explore--Deepen--Validate--Synthesize paradigm.

\textbf{Layer 2 --- Utility-Aligned Profile Exploration (offline).} For each cluster $a\in\mathcal{A}$, the LLM samples candidate profiles conditioned on $\mathcal{K}$ and the aggregated cluster history, ranks them via a lightweight LOGIC-rule proxy of downstream prediction utility, iteratively refines the best, and constructs preference pairs that drive a one-time DPO fine-tune of the profile generator. The resulting DPO-aligned generator emits a single profile per cluster, encoded into a $d$-dimensional embedding $\mathbf{e}_a$ by a frozen text encoder.

\textbf{Layer 3 --- Online Outcome Prediction \& Matching.} The latency-critical dispatcher performs only two operations per OD pair: (i) a deterministic cluster-assignment rule evaluation against $\mathcal{A}$ to look up the user's cluster, and (ii) a cached embedding fetch returning $\mathbf{e}_p$ and $\mathbf{e}_d$. These embeddings are concatenated with the existing structured-feature vector and fed to the production multi-task predictor; matching weights and KM assignment proceed unchanged. \emph{No LLM inference occurs on this path}, and combined overhead is under $0.01$\,ms per pair.

\textbf{The offline--online contract.} Layers~1--2 emit two and only two artifacts that cross into Layer~3: the cluster-assignment rule set $\mathcal{A}$ (a few KB of interpretable Boolean rules) and the cluster-embedding table $\{\mathbf{e}_a\}$. This narrow interface is the structural reason ProfiLLM operates within DiDi's existing latency budget without modifying the downstream matching components.

\subsection{Tool-Augmented Global Knowledge Mining}
\label{subsec:knowledge_mining}

Directly feeding massive historical logs to LLMs is infeasible due to context length limits and computational costs. 
To address Challenge 1, we design a tool-augmented knowledge mining module that enables an LLM agent to systematically analyze platform-scale data through structured tool invocations.

\subsubsection{Tool Design and Categorization}
\label{subsubsec:tools}

We design a comprehensive toolkit $\mathcal{T}$ containing 27 analytical tools organized into six categories.
Each tool $t_i \in \mathcal{T}$ is defined by a tuple $(\textit{name}_i,\allowbreak \textit{desc}_i,\allowbreak \textit{params}_i,\allowbreak \textit{func}_i)$, where $\textit{name}_i$ is the tool identifier, $\textit{desc}_i$ provides a natural language description for the LLM to understand its functionality, $\textit{params}_i$ specifies the required input parameters, and $\textit{func}_i$ implements the actual computation logic. The tools are designed to be composable, allowing the agent to chain multiple tools to answer complex analytical questions.

\subsubsection{Explore-Deepen-Validate-Synthesize Paradigm}
\label{subsubsec:edvs}

We structure the knowledge mining process into four phases that guide the LLM agent through systematic data exploration:
\textbf{(1) Explore:} The agent invokes basic statistical tools to compute platform-level metrics (\eg completion rate, acceptance rate), examine feature distributions, and surface preliminary patterns.
\textbf{(2) Deepen:} Based on exploration findings, the agent conducts focused analyses on promising directions, including temporal patterns (\eg hourly cancellation variations), spatial heterogeneity (\eg grid-level supply-demand imbalances), and user segmentation (\eg clustering by historical order patterns).
\textbf{(3) Validate:} Discovered patterns are subjected to statistical hypothesis testing and causal analysis; only findings with significant effect sizes and passing appropriate tests are retained.
\textbf{(4) Synthesize:} Validated findings are consolidated into three structured outputs: \textit{Global Knowledge $\mathcal{K}$}, containing platform benchmarks, temporal regularities, weather impacts, and causal relationships; \textit{User Clustering Rules $\mathcal{A}$}, where each cluster $a \in \mathcal{A}$ is defined by a rule-based classifier $\phi_a: \mathcal{H}_u \rightarrow \{0, 1\}$ over user history (\eg active hours, frequent regions, cancellation patterns); and \textit{Regional Priors $\mathcal{R}$}, storing grid-level and time-slot-level supply-demand statistics $P(\textit{supply}, \textit{demand} \mid g, s)$ with derived metrics such as expected waiting time and fulfillment rate.

\subsection{Utility-Aligned Profile Exploration}
\label{subsec:dpo_profile}

With the mined global knowledge and clustering rules, we now address Challenges 2 and 3 through a cluster-level profile exploration mechanism. 
The key insight is that: (1) clustering aggregates sufficient historical data for reliable profiling even for low-frequency users, and (2) profile quality should be measured by downstream prediction performance rather than surface-level fluency.

\subsubsection{User Clustering}
\label{subsubsec:clustering}

Using the clustering rules $\mathcal{A}$ from the knowledge mining module, we partition all users into $|\mathcal{A}|$ clusters. For each user $u \in \mathcal{P} \cup \mathcal{D}$, we assign them to the most appropriate cluster based on their historical behavior: $a^*(u) = \arg\max_{a \in \mathcal{A}} \phi_a(\mathcal{H}_u)$, where $\phi_a(\mathcal{H}_u) \in [0, 1]$ evaluates the compatibility between user $u$'s history and cluster $a$'s defining characteristics. Users within the same cluster share similar behavioral patterns, allowing us to learn a shared user profile that captures cluster-level regularities.

Let $\mathcal{U}_a = \{u : a^*(u) = a\}$ denote the set of users assigned to cluster $a$, and $\mathcal{H}_a = \bigcup_{u \in \mathcal{U}_a} \mathcal{H}_u$ denote the aggregated historical orders for the cluster. This aggregation ensures sufficient data volume for reliable profile learning, addressing Challenge 2.

\subsubsection{Structured Profile Generation}
\label{subsubsec:candidate}

For each cluster $a \in \mathcal{A}$, we prompt the LLM to generate $K$ diverse candidate user profiles based on the aggregated cluster history $\mathcal{H}_a$ and global knowledge $\mathcal{K}$. Each candidate profile follows a structured three-part format:
\begin{equation}
    \{\textit{profile}_a^{(k)}\}_{k=1}^{K} = \text{LLM}(\mathcal{H}_a, \mathcal{K}, \textit{prompt}_{\textit{gen}})
\end{equation}
Each profile $\textit{profile}_a^{(k)}$ consists of three components:
(1) \textbf{ANALYSIS}: A detailed examination of the cluster's behavioral patterns observed in the historical data, identifying key factors that influence order outcomes.
(2) \textbf{PROFILE}: A semantic description characterizing the cluster's order completion patterns, behavioral preferences, and service compatibility signals (\eg efficiency-oriented drivers who target high earnings-per-effort rides).
(3) \textbf{LOGIC}: Executable decision rules derived from the analysis and profile, expressed as Boolean conditions over order features (\eg \texttt{(Price\_Per\_KM >= 2.5) AND (Pick\_KM <= 2.0)}).
The prompt $\textit{prompt}_{\textit{gen}}$ instructs the LLM to generate diverse profiles emphasizing different aspects (\eg temporal patterns, price sensitivity, spatial preferences) while remaining grounded in the cluster's actual order history.

\subsubsection{Utility-Based Profile Evaluation}
\label{subsubsec:evaluation}

A critical design choice is how to evaluate profile quality efficiently. Rather than training full prediction models for each candidate, which would be computationally prohibitive, we leverage the LOGIC component as a lightweight proxy for utility evaluation. The intuition is that if the rules extracted from a profile accurately predict outcomes, then the profile captures meaningful behavioral patterns that are also likely to benefit the embedding-based predictor.

Specifically, for each candidate profile $\textit{profile}_a^{(k)}$, we extract its LOGIC rules and apply them to the cluster's historical order-driver pairs $\{(c_i, y_i)\}_{i=1}^{n_a}$ to generate rule-based predictions $\{\hat{y}_i^{(k)}\}$:
\begin{equation}
    \hat{y}_i^{(k)} = \texttt{evaluate}(\textit{LOGIC}_a^{(k)}, c_i).
\end{equation}

We then compute a fused prediction by blending the base model's output with the LOGIC-rule prediction. Specifically, for each order-driver pair $c_i$ in cluster $a$ and each prediction task $l$ (\eg driver cancellation), the base production model produces a probability estimate $\hat{y}_{i,\text{base}}^{(l)} \in [0,1]$, while the LOGIC rules yield a binary prediction $\hat{y}_{i,\text{logic}}^{(k)} \in \{0, 1\}$. The fused prediction is obtained via a convex combination controlled by a blending coefficient $\lambda$:
\begin{equation}
    \hat{y}_{i,\text{fused}}^{(k)} = (1 - \lambda) \cdot \hat{y}_{i,\text{base}}^{(l)} + \lambda \cdot \hat{y}_{i,\text{logic}}^{(k)}
    \label{eq:fused}
\end{equation}
where $\lambda \in [0, 1]$ governs the relative influence of the LOGIC rules. The fused AUC is then computed against the ground-truth labels $\{y_i^{(l)}\}_{i=1}^{n_a}$:
\begin{equation}
    \text{AUC}_a^{(k)} = \text{ComputeAUC}\left(\{y_i^{(l)}\}_{i=1}^{n_a},\; \{\hat{y}_{i,\text{fused}}^{(k)}\}_{i=1}^{n_a}\right).
    \label{eq:utility_gain}
\end{equation}
The utility gain of profile $k$ is measured as the AUC improvement over the base production model alone: $\Delta_a^{(k)} = \text{AUC}_a^{(k)} - \text{AUC}_a^{\text{base}}$, where $\text{AUC}_a^{\text{base}} = \text{ComputeAUC}(\{y_i^{(l)}\},\, \{\hat{y}_{i,\text{base}}^{(l)}\})$ is AUC achieved by the base model without LOGIC-rule augmentation. A positive $\Delta_a^{(k)}$ indicates that the LOGIC rules provide complementary signals that improve prediction beyond the existing production model.

This fused evaluation serves as an efficient and informative proxy for profile quality. Since the base model already captures patterns available from structured features, a positive $\Delta_a^{(k)}$ directly indicates that the LOGIC rules encode complementary behavioral signals invisible to the production predictor. Moreover, the blending mechanism ensures that only profiles contributing genuine discriminative power beyond the existing model are favored, filtering out superficially plausible descriptions that merely recapitulate what structured features already capture, thereby reliably identifying profiles most beneficial for downstream outcome prediction.

\subsubsection{Iterative Profile Refinement}
\label{subsubsec:refinement}

We adopt an iterative refinement strategy to progressively improve profile quality. Starting from the best-performing initial candidate $\textit{profile}_a^{(k^*)}$ where $k^* = \arg\max_k \Delta_a^{(k)}$, we prompt the LLM to generate refined versions that address identified weaknesses:
\begin{equation}
    \textit{profile}_a^{(t+1)} = \text{LLM}(\textit{profile}_a^{(t)}, \mathcal{H}_a, \textit{feedback}^{(t)}, \textit{prompt}_{\textit{refine}})
\end{equation}
where $\textit{feedback}^{(t)}$ summarizes the prediction errors made by the current profile's LOGIC rules (\eg false positive/negative cases with their order features). The refinement continues for $T$ iterations or until the utility gain plateaus, yielding the selected profile $\textit{profile}_a^* = \arg\max_{t \in \{1, \ldots, T\}} \Delta_a^{(t)}$.


\subsubsection{DPO Fine-tuning for Profile Generation}
\label{subsubsec:dpo_finetune}

The exploration process naturally produces preference pairs that can be leveraged for DPO fine-tuning. For each cluster $a$, we construct preference pairs by comparing profiles with different utility gains:
\begin{equation}
    \mathcal{P}_a = \{(\mathcal{H}_a, \textit{profile}_w, \textit{profile}_l) : \Delta_a^{(w)} > \Delta_a^{(l)} + \gamma\}
\end{equation}
where $\gamma$ is a margin threshold ensuring meaningful preference differences. Aggregating across all clusters yields a preference dataset $\mathcal{P} = \bigcup_{a \in \mathcal{A}} \mathcal{P}_a$.
We perform a one-time offline DPO fine-tuning of the LLM using the standard DPO objective~\cite{rafailov2023direct} on $\mathcal{P}$. This fine-tuning aligns the LLM's profile generation capability with downstream prediction utility, improving the quality of profiles for newly defined clusters without requiring additional exploration iterations.

Note that all profiles describe cluster-level behavioral patterns (\eg temporal preferences, price sensitivity, cancellation tendencies) rather than individual demographic attributes, ensuring that the profiling mechanism operates on aggregate behavioral signals without raising privacy concerns.

\subsection{Online Outcome Prediction and Matching}
\label{subsec:prediction}


\subsubsection{Profile Embedding and Caching}
\label{subsubsec:embedding}

Each cluster's textual PROFILE component is converted into a dense embedding via a pre-trained text encoder $\mathcal{E}$:$\mathbf{e}_a = \mathcal{E}(\textit{PROFILE}_a^*) \in \mathbb{R}^d$.
For each user $u$, their profile embedding is simply the embedding of their assigned cluster: $\mathbf{e}_u = \mathbf{e}_{a^*(u)}$. Since profiles are at the cluster level, the number of embeddings to cache is bounded by $|\mathcal{A}|$, making caching highly efficient. All cluster embeddings are pre-computed and stored in a distributed cache with sub-millisecond lookup latency.

\subsubsection{Online Feature Integration and Matching}
\label{subsubsec:integration}

At serving time, for each candidate order-driver pair $c = (o, d)$ with passenger $p$, we construct the input representation by concatenating structured features with the cached profile embeddings: $\mathbf{z}_c = [\mathbf{x}_c; \mathbf{e}_p; \mathbf{e}_d; \mathbf{r}_{g,s}]$, where $\mathbf{x}_c$ denotes the original structured features, $\mathbf{e}_p$ and $\mathbf{e}_d$ are the profile embeddings retrieved from cache based on user-to-cluster assignments, and $\mathbf{r}_{g,s} \in \mathcal{R}$ is the regional supply-demand prior for grid $g$ and time slot $s$.
The multi-task prediction model $f_\theta$ jointly predicts order-level outcomes $\hat{\mathbf{y}}_c = f_\theta(\mathbf{z}_c)$, which are then aggregated into a matching weight $w_c$ quantifying the expected utility of each assignment. The regional prior $\mathcal{R}$ is incorporated to adjust weights based on supply-demand conditions. Finally, the optimal matching $\mathcal{M}^*$ that maximizes $\sum_{c \in \mathcal{M}} w_c$ is computed via the Kuhn-Munkres algorithm, completing one dispatching cycle.

Since no LLM inference occurs at serving time and user-to-cluster assignment is a simple rule evaluation, the latency from profile is negligible. For new users without sufficient history, the system assigns them to a default cluster, ensuring consistent service quality.

\begin{table*}[t]
\renewcommand{\arraystretch}{0.3}
\setlength{\tabcolsep}{12pt}
\small
\centering
\caption{Relative improvement (\%) over pickup-distance-based KM matching across three cities and time periods.%
\textsuperscript{\,\dag}\,Time periods: Morning = 7:00--10:00, Noon = 11:00--14:00, Evening = 17:00--20:00.}
\label{tab:results}
\vspace{-0.3cm}
\begin{tabular}{@{} cl l|cc|cc|cc|cc @{}}
\toprule
\multirow{2}{*}{\textbf{City}} 
& \multirow{2}{*}{\textbf{Group}} 
& \multirow{2}{*}{\textbf{Method}} 
& \multicolumn{2}{c|}{\textbf{Overall}} 
& \multicolumn{2}{c|}{\textbf{Morning}\textsuperscript{\dag}} 
& \multicolumn{2}{c|}{\textbf{Noon}\textsuperscript{\dag}} 
& \multicolumn{2}{c}{\textbf{Evening}\textsuperscript{\dag}} \\ 
\cmidrule(lr){4-5} \cmidrule(lr){6-7} \cmidrule(lr){8-9} \cmidrule(lr){10-11}
& & & \textbf{GMV} & \textbf{CR} & \textbf{GMV} & \textbf{CR} & \textbf{GMV} & \textbf{CR} & \textbf{GMV} & \textbf{CR} \\ 
\midrule

\multirow{12}{*}{\rotatebox{0}{\textbf{City A}}} 
& \multirow{2}{*}{\textit{Trad.}}
  & TVal                  & +2.24 & +2.14  & +3.86  & +2.44  & +1.70  & +1.08  & +2.06  & +1.41 \\
& & GRC                   & +0.73 & \negv{$-$3.42}  & +4.18  & +0.62  & +3.01  & +0.22  & \negv{$-$2.36}  & \negv{$-$3.97} \\

\cmidrule(lr){2-11}

& \multirow{7}{*}{\textit{LLM}}
  & Llama-3.3-70B         & +2.34 & +2.76  & +0.84  & +1.49  & +1.02  & +3.00  & +2.21  & +7.12 \\
& & Qwen3-Next-80B        & +2.41 & +2.54  & +1.58  & +1.54  & +1.80  & +3.89  & +2.80  & +6.29 \\
& & DeepSeek-R1           & +2.53 & +4.57  & +1.40  & +2.17  & +1.35  & +3.62  & +2.60  & +7.61 \\
& & Kimi-K2               & +1.96 & +4.77  & +2.04  & +1.78  & +0.93  & +3.34  & +2.43  & +8.15 \\
& & GPT-OSS-120B          & +2.44 & +5.75  & +0.75  & +1.65  & +1.95  & +4.62  & +0.05  & +6.96 \\
& & Gemini-3-Flash        & +1.41 & +4.62  & +0.34  & +1.01  & +1.57  & +4.20  & \negv{$-$0.26}  & +5.63 \\
& & Gemini-3-Pro          & +2.95 & +5.48  & +3.96  & +4.72  & +1.74  & +3.50  & +1.62  & +5.93 \\

\cmidrule(lr){2-11}

& \multirow{2}{*}{\textit{Ours}}
  & \textbf{ProfiLLM-DPO}   & \best{+4.02} & {+6.03}  & {+4.97}  & {+5.14}  & {+3.01}  & {+0.22}  & {+2.82}  & \best{+9.94} \\
&  & \textbf{ProfiLLM}         & {+3.52} & \best{+7.10}  & \best{+5.02}  & \best{+5.67}  & \best{+3.20}  & \best{+4.98}  & \best{+3.19}  & {+6.97} \\

\midrule
\addlinespace[2pt]

\multirow{12}{*}{\rotatebox{0}{\textbf{City B}}} 
& \multirow{2}{*}{\textit{Trad.}}
  & TVal                  & +1.87 & +1.63  & +3.12  & +1.98  & +1.24  & +0.76  & +1.72  & +1.05 \\
& & GRC                   & +1.15 & \negv{$-$2.18}  & +3.54  & +0.38  & +2.47  & \negv{$-$0.15}  & \negv{$-$1.82}  & \negv{$-$3.25} \\

\cmidrule(lr){2-11}

& \multirow{7}{*}{\textit{LLM}}
  & Llama-3.3-70B         & +1.92 & +2.31  & +0.56  & +1.12  & +0.78  & +2.54  & +1.87  & +6.38 \\
& & Qwen3-Next-80B        & +2.08 & +2.12  & +1.24  & +1.18  & +1.45  & +3.25  & +2.42  & +5.67 \\
& & DeepSeek-R1           & +2.17 & +3.89  & +1.08  & +1.82  & +1.02  & +3.08  & +2.24  & +6.83 \\
& & Kimi-K2               & +1.63 & +4.05  & +1.72  & +1.45  & +0.68  & +2.87  & +2.08  & +7.34 \\
& & GPT-OSS-120B          & +2.06 & +5.12  & +0.48  & +1.32  & +1.62  & +4.08  & \negv{$-$0.18}  & +6.24 \\
& & Gemini-3-Flash        & +1.08 & +3.94  & +0.12  & +0.78  & +1.24  & +3.65  & \negv{$-$0.48}  & +4.89 \\
& & Gemini-3-Pro          & +2.51 & {+4.83}  & +3.42  & +4.15  & +1.38  & +2.94  & +1.28  & +5.18 \\

\cmidrule(lr){2-11}

& \multirow{2}{*}{\textit{Ours}}
  & \textbf{ProfiLLM-DPO}   & \best{+3.58} & {+5.47}  & +4.42  & {+4.68}  & {+2.64}  & \negv{$-$0.08}  & {+2.48}  & \best{+9.12} \\
&  & \textbf{ProfiLLM}      & {+3.14} & \best{+6.52}  & \best{+4.56}  & \best{+5.24}  & \best{+2.82}  & \best{+4.35}  & \best{+2.84}  & +6.28 \\

\midrule
\addlinespace[2pt]

\multirow{12}{*}{\rotatebox{0}{\textbf{City C}}} 
& \multirow{2}{*}{\textit{Trad.}}
  & TVal                  & +2.56 & +2.48  & +4.24  & +2.82  & +1.92  & +1.34  & +2.38  & +1.72 \\
& & GRC                   & +0.41 & \negv{$-$1.87}  & +2.86  & +0.54  & +1.78  & +0.08  & \negv{$-$1.54}  & \negv{$-$2.83} \\

\cmidrule(lr){2-11}

& \multirow{7}{*}{\textit{LLM}}
  & Llama-3.3-70B         & +2.68 & +3.12  & +1.12  & +1.78  & +1.34  & +3.42  & +2.54  & +7.56 \\
& & Qwen3-Next-80B        & +2.75 & +2.89  & +1.82  & +1.76  & +2.04  & +4.12  & +3.12  & +6.72 \\
& & DeepSeek-R1           & +2.91 & +4.93  & +1.64  & +2.48  & +1.58  & +3.94  & +2.92  & +7.98 \\
& & Kimi-K2               & +2.24 & +5.18  & +2.36  & +2.04  & +1.12  & +3.68  & +2.72  & +8.52 \\
& & GPT-OSS-120B          & +2.79 & +6.08  & +0.98  & +1.86  & +2.18  & +4.84  & +0.32  & +7.38 \\
& & Gemini-3-Flash        & +1.72 & +4.95  & +0.58  & +1.24  & +1.82  & +4.48  & +0.08  & +5.42 \\
& & Gemini-3-Pro          & +3.28 & {+5.81}  & {+4.28}  & {+5.04}  & +1.98  & +3.72  & +1.92  & +6.28 \\

\cmidrule(lr){2-11}

& \multirow{2}{*}{\textit{Ours}}
  & \textbf{ProfiLLM-DPO}   & \best{+4.35} & +6.41  & +5.24  & +5.48  & +3.38  & +0.48  & +3.14  & \best{+10.32} \\
&  & \textbf{ProfiLLM}         & +3.87 & \best{+7.53}  & \best{+5.38}  & \best{+5.92}  & \best{+3.54}  & \best{+5.24}  & \best{+3.48}  & +7.42 \\

\bottomrule
\end{tabular}
\vspace{-0.3cm}
\end{table*}


\section{Experiments}
\label{sec:experiments}

We conduct extensive experiments on real-world industrial datasets from DiDi's ride-hailing platform to evaluate ProfiLLM.
Due to space, extended analyses: the dispatching simulator design, discovered cluster archetypes, cluster-count and $\lambda$ sensitivity, offline cost and complexity, the 14-day stability study, prompt templates, are deferred to our online appendix.\footnote{\url{https://ProfiLLM.github.io}}

\subsection{Experimental Setup}
\label{subsec:setup}

\subsubsection{Datasets}
We evaluate ProfiLLM on real-world order dispatching from three Brazilian cities spanning distinct supply--demand regimes: \textbf{City~A} (medium-scale, supply-constrained, highest driver utilization), \textbf{City~B} (medium-scale, supply-relaxed, moderate order density), and \textbf{City~C} (large-scale, high-demand, complex traffic).
%
For each city, we use 38 days of historical data for training (including global knowledge mining and profile exploration) and five days for testing. 
The City~A analysis window covers 333{,}166 active passengers and 12{,}128 active drivers, with the long-tail and heterogeneity statistics applying directly to this evaluation set.

\subsubsection{Evaluation Metrics}
\label{subsubsec:metrics}
We evaluate performance at two levels.
\textit{(i) Dispatching level (realized rates).}
We report \textbf{GMV} (Gross Merchandise Value) together with six realized rates computed in the dispatching simulator or the online A/B test:
\textbf{CR} (Completion Rate),
\textbf{DAR} (Driver Acceptance Rate),
\textbf{DCR} (Driver Cancellation Rate, post-acceptance driver cancel),
\textbf{PCR} (Passenger Cancellation Rate, post-acceptance passenger cancel),
\textbf{CBA} (Cancel Before Accept, share of orders the passenger cancels before any driver accepts), and
\textbf{BER} (Bad Experience Rate, share of completed orders with excessively long pickup distance).
\textit{(ii) Prediction level (AUC).}
For each candidate OD pair, the prediction model produces a probability for each of four behavioral events. We report \textbf{AUC} for each task, denoted \textbf{Accept}, \textbf{D-Cancel}, \textbf{P-Cancel}, and \textbf{Success}, in one-to-one correspondence with DAR, DCR, PCR, and CR.
The two views measure the same four events at different stages of the pipeline and can therefore move differently when a system change affects matching weights more than per-pair prediction (or vice versa).
CBA and BER are pure dispatching-level rates without a direct prediction-task counterpart, since CBA occurs before any OD-pair assignment is made and BER reflects realized pickup quality after matching.

\subsubsection{Baselines}

We compare ProfiLLM against two categories of baselines. \textbf{Traditional dispatching methods} rely on structured features without user profiling: TVal~\cite{xu2018large} and GRC~\cite{yang2024rethinking}. \textbf{LLM-based profiling methods} use the same clustering and knowledge mining pipeline as ProfiLLM but differ in the LLM backbone for profile generation, including Llama-3.3-70B~\cite{dubey2024llama}, Qwen3-Next-80B~\cite{yang2025qwen3}, DeepSeek-R1~\cite{guo2025deepseek}, Kimi-K2~\cite{team2025kimi}, Gemini-3-Flash~\cite{team2023gemini}, Gemini-3-Pro~\cite{team2023gemini}, and GPT-OSS-120B. \textbf{ProfiLLM} with exploration-selected profiles but without DPO fine-tuning, and \textbf{ProfiLLM-DPO} with the full framework including DPO-aligned profile generation.

\subsubsection{Implementation Details}
For the global knowledge mining, we use Gemini-3-Pro as the backbone LLM agent with a temperature of 0.3 for consistent tool invocation. 
For profile exploration, we generate $K=5$ initial candidate profiles per cluster and perform $T=3$ refinement iterations. The margin threshold $\gamma$ for constructing DPO preference pairs is set to 0.001 AUC improvement. We fine-tune Qwen3-8B via DPO on the collected preference pairs as the aligned profile generator.
The text encoder $\mathcal{E}$ for embedding user profiles is a fine-tuned sentence transformer with output dimension $d=768$. All offline experiments are conducted on a cluster with 8 NVIDIA L20 GPUs. Dispatching quality is evaluated with a replay-based simulator that re-runs historical orders and driver availability per city in production-identical 2-second Kuhn--Munkres cycles, sampling Accept/Cancel outcomes from the multi-task predictor. Online serving operates on pre-computed cluster embeddings stored in a distributed cache, ensuring sub-millisecond lookup latency with negligible overhead to the existing prediction pipeline.

\begin{table*}[t]
\renewcommand{\arraystretch}{0.3}
\setlength{\tabcolsep}{3.8pt}
\small
\centering
\caption{Multi-task prediction AUC improvement (\%) over Structured Only baseline. Higher is better. Best results are in \textbf{bold}.}
\vspace{-0.4cm}
\begin{tabular}{@{}l|cccc|cccc|cccc@{}}
\toprule
\multirow{2}{*}{\textbf{Method}} & \multicolumn{4}{c|}{\textbf{City A}} & \multicolumn{4}{c|}{\textbf{City B}} & \multicolumn{4}{c}{\textbf{City C}} \\
\cmidrule(lr){2-5} \cmidrule(lr){6-9} \cmidrule(lr){10-13}
& \textbf{Accept} & \textbf{D-Cancel} & \textbf{P-Cancel} & \textbf{Success} & \textbf{Accept} & \textbf{D-Cancel} & \textbf{P-Cancel} & \textbf{Success} & \textbf{Accept} & \textbf{D-Cancel} & \textbf{P-Cancel} & \textbf{Success} \\
\midrule
Llama-3.3-70B       & -1.10 & -0.71	& +0.19 &	-1.14 & -0.64 & +0.38 & -0.38 & -0.45 & -0.01 & -0.34 & +0.25 & +0.01  \\
Qwen3-Next-80B      & -0.22 &	-0.38 & +1.65 &	+0.02 & -0.52 & -0.40 & -5.71 & -7.57 & -0.03 & -0.16 & +0.27 & -0.06 \\
DeepSeek-R1         & +0.06 & +0.23 & +2.05 & +0.25 & +0.31 & +1.85 & +1.06 & +0.48 & +0.21 & -0.13 & +0.04 & +0.14 \\
Kimi-K2             & -0.17 & +0.82	& +2.11 & -0.07 & -2.44 & -0.44 & -6.33 & -1.91 & +0.50 & -0.11 & +0.40 & +0.45 \\
Gemini-3-Flash      & +0.10 & +0.53 & +1.83 & +0.42 & +0.24 & +1.76 & -0.11 & +0.38 & +0.03 & -0.26 & +0.37 & +0.04 \\
Gemini-3-Pro        & -0.08 & -0.68 & +2.37 & +0.56 & -0.44 & +0.50	& +0.24 & -0.31 & +0.02 & -0.03 & +0.10 & +0.05 \\
GPT-OSS-120B        & -0.02 & +0.14 & +1.83 & +0.17 & +0.11 & +1.64	& +0.63 & +0.29 & -0.09 & -0.02 & +0.44 & -0.06 \\
\midrule
\textbf{ProfiLLM-DPO}          & +1.51 & +2.76 & +6.02 & +1.72 & +2.25 & +4.98	& +5.55 & +2.58 & +0.65 & +5.93 & +5.30 & +2.37 \\
\textbf{ProfiLLM}       & +1.56 & +3.88	& +6.14 & +1.80 & +2.26 & +4.98	& +6.00 & +2.60 & +0.84 & +5.95 & +5.65 & +2.48 \\
\bottomrule
\end{tabular}
\label{tab:auc_results}
\vspace{-0.3cm}
\end{table*}

\subsection{Overall Performance}
\label{subsec:overall}

Table~\ref{tab:results} presents the simulator-based evaluation results across three cities. 
%
From the results, we make the following observations.
(1) ProfiLLM variants consistently outperform all baselines. Both ProfiLLM and ProfiLLM-DPO achieve the highest GMV and CR among all methods. Notably, the two variants exhibit complementary strengths: ProfiLLM-DPO achieves the best GMV (\eg +4.35\% in City C), while ProfiLLM attains the best CR (\eg +7.53\% in City C). This suggests that DPO fine-tuning steers profiles toward revenue-critical signals at a slight cost to CR, whereas exploration-selected profiles maintain a more balanced optimization across the fulfillment funnel.
(2) LLM-based profiles substantially outperform traditional methods. TVal and GRC, which rely solely on structured numerical features, achieve moderate GMV gains but show inconsistent CR performance (\eg GRC yields negative CR in all three cities). In contrast, all LLM-based methods deliver positive improvements on both metrics, confirming that semantic user profiles capture behavioral patterns invisible to handcrafted features.
(3) Utility alignment matters more than model scale. Among LLM-based methods, larger backbones generally yield better profiles, yet our DPO-aligned Qwen3-8B consistently outperforms much larger models such as Gemini-3-Pro without alignment. This demonstrates that optimizing for downstream prediction utility is more effective than scaling the LLM backbone alone.

\begin{figure}[h]
\centering
\includegraphics[width=0.9\linewidth]{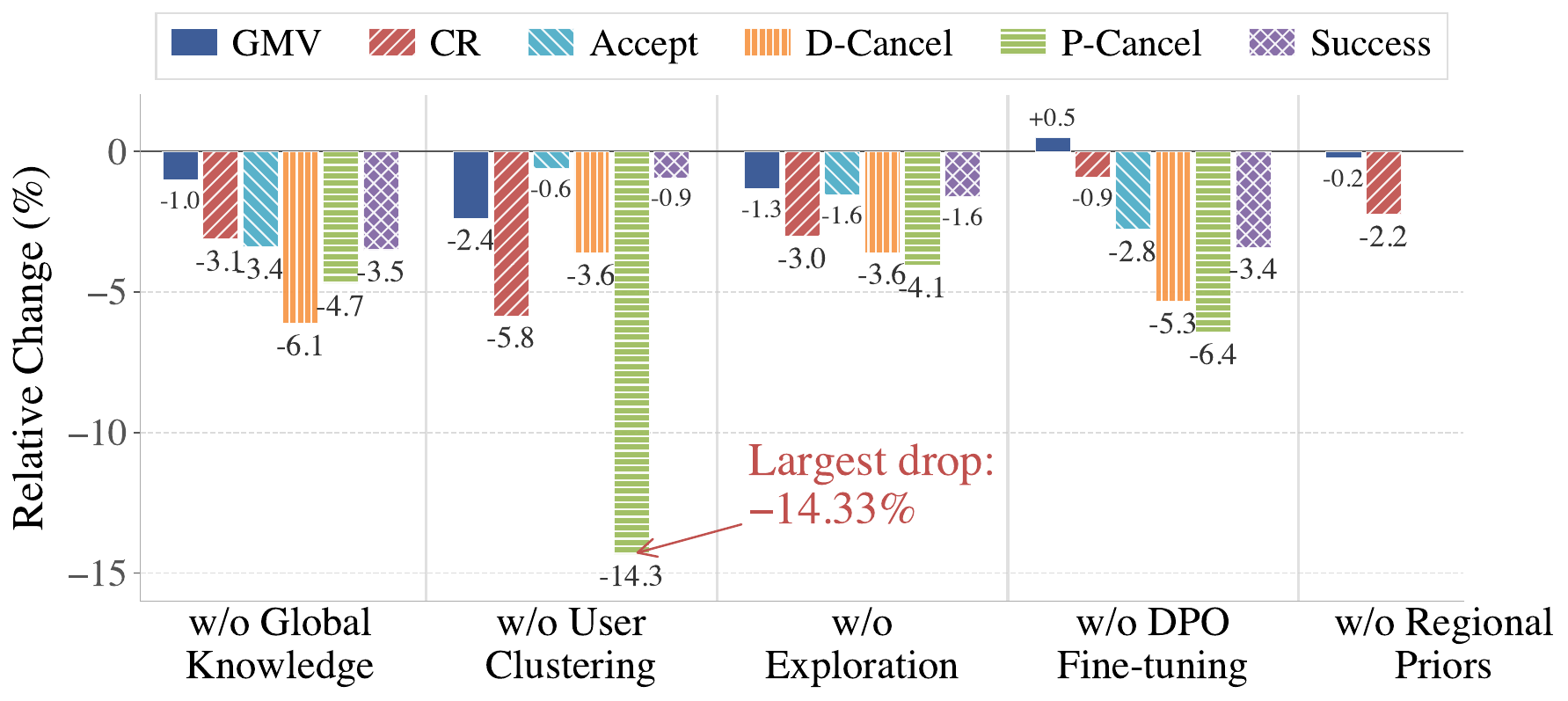}
\vspace{-0.4cm}
\caption{Ablation on City~A. \emph{CR} is the simulator's realized completion rate; \emph{Success} is the per-OD-pair completion AUC. Regional Priors only affect matching weights, hence no prediction-AUC bar.}
\Description{A grouped bar chart showing the relative change in dispatching and prediction metrics when each ProfiLLM component is removed. Bars cover CR, Success, and the cancellation AUC tasks, with all variants degrading relative to the full model; removing User Clustering or Global Knowledge produces the largest drops, while removing Regional Priors mainly affects dispatching CR and is not applicable for prediction AUC.}
\label{fig:ablation}
\vspace{-0.5cm}
\end{figure}

\subsection{Prediction Performance}
\label{subsec:exp_prediction}
To evaluate the quality of LLM-generated profiles for outcome prediction, we compare the AUC of multi-task prediction models trained with profiles from different methods. Table~\ref{tab:auc_results} reports the AUC improvement over the structured-only baseline for prediction tasks.
The results reveal that naively applying LLMs for profile generation does not guarantee downstream utility. While some baseline LLMs achieve modest positive improvements, others actually degrade prediction performance: Kimi-K2 drops P-Cancel AUC by 6.33\% in City B, and Qwen3-Next-80B yields a 7.57\% decrease in Success. In contrast, both ProfiLLM and ProfiLLM-DPO deliver consistent positive gains across all tasks and cities, with ProfiLLM achieving up to +6.14\% on P-Cancel (City A), +5.95\% on D-Cancel (City C), and +2.60\% on Success (City B). The largest improvements are observed for cancellation prediction, which aligns with our pilot study: cancellation behaviors depend heavily on contextual preferences such as sensitivity to pickup distance that handcrafted features struggle to capture. Finally, ProfiLLM and ProfiLLM-DPO achieve comparable prediction AUC, with ProfiLLM slightly ahead on most metrics, suggesting that iterative exploration is the primary driver of prediction gains, while DPO fine-tuning trades a small amount of per-cluster local optimality for cross-cluster generalization and substantially lower offline refresh cost.

\begin{figure}[t]
\centering
\includegraphics[width=0.8\linewidth]{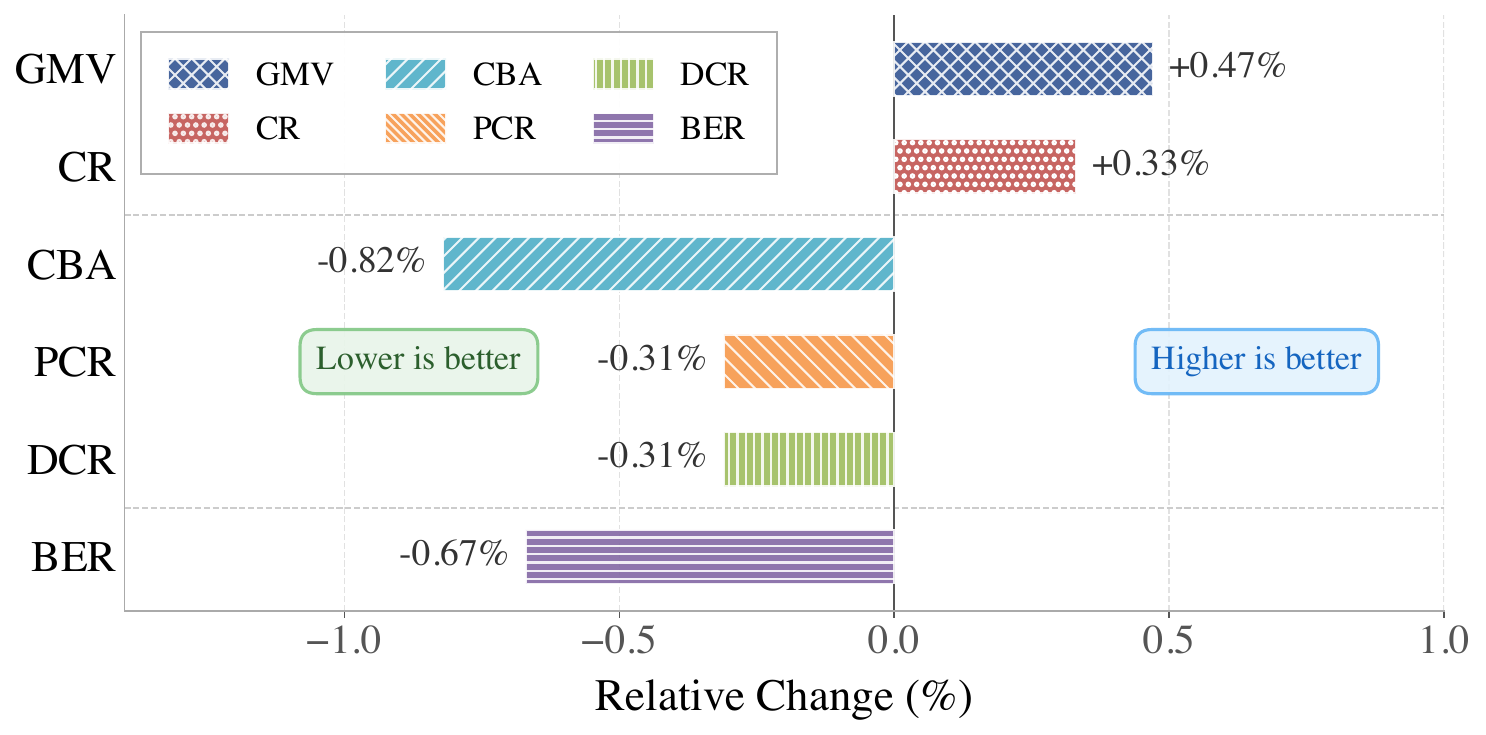}
\vspace{-0.4cm}
\caption{Online A/B over 14 days in City~A: treatment-vs.-control relative change.}
\Description{A bar chart of relative treatment-versus-control changes for six production metrics from a 14-day online A/B test in City~A, grouped by funnel stage. GMV and CR rise, while CBA, PCR, DCR, and BER fall, with each bar moving in the operationally desirable direction.}
\label{fig:online_ab}
\vspace{-0.6cm}
\end{figure}

\subsection{Ablation Study}
\label{subsec:ablation}
Figure~\ref{fig:ablation} reports results on City A as relative change compared to the full ProfiLLM.
(1) \textbf{w/o Global Knowledge:} Removing global knowledge causes substantial drops across all tasks (\eg D-Cancel $-$6.12\%), confirming that platform-level knowledge provides essential grounding for generating predictive profiles.
(2) \textbf{w/o User Clustering:} Replacing cluster-level profiles with per-user profiles yields the largest degradation (P-Cancel $-$14.33\%), as 96\% of passengers lack sufficient history for reliable individual profiling.
(3) \textbf{w/o Exploration:} Skipping iterative refinement degrades both dispatching (GMV $-$1.32\%) and prediction, confirming that the explore-evaluate-refine loop is necessary for utility alignment.
(4) \textbf{w/o DPO Fine-tuning:} We replace the DPO-aligned generator with the un-finetuned base Qwen3-8B in a single-pass setting (no exploration, no alignment). Prediction AUC drops substantially (P-Cancel $-$6.42\%), confirming that preference alignment is what makes the compact 8B generator competitive. Note that this ablation differs from the ProfiLLM variant in Table~\ref{tab:auc_results}, which uses Gemini-3-Pro with exploration but no DPO; that path achieves comparable AUC to ProfiLLM-DPO at higher offline LLM cost.
(5) \textbf{w/o Regional Priors:} Primarily impacts dispatching (CR $-$2.24\%) without affecting prediction, consistent with priors being incorporated into matching weights rather than the prediction model.

\subsection{Production Deployment}
\label{subsec:deployment}
We deployed ProfiLLM in DiDi's production environment and report results from a 14-day A/B test in City~A (extending the initial 5-day pilot to a longer window for more stable estimates).
As shown in Figure~\ref{fig:online_ab}, ProfiLLM achieves consistent improvements across every monitored realized rate (see Section~\ref{subsubsec:metrics} for definitions). Revenue and completion improve simultaneously: GMV rises by 0.47\% and CR by 0.33\%, confirming that the gains stem from higher matching quality rather than from trading off completion for revenue. Post-acceptance cancellations decrease on both sides of the match: PCR drops by 0.31\% and DCR by 0.31\%. The two non-prediction funnel rates also move in the desired direction: CBA drops by 0.82\%, indicating fewer passengers walk away before a driver accepts, and BER drops by 0.67\%, indicating fewer completed orders end up with poor pickup quality. The simultaneous beneficial movement across mechanistically distinct funnel stages provides converging evidence that ProfiLLM systematically improves the dispatching components it was designed to enhance.
These online deltas are several times smaller than the simulation gains of Section~\ref{subsec:overall} (\eg +0.47\% vs.\ +4.02\% GMV in City~A); this gap is expected by construction, so we read the simulator for effect \emph{sign} and method \emph{ranking} rather than absolute magnitude (validated in our online appendix).

\begin{table}[t]
\centering
\caption{Offline pipeline cost breakdown in City~A. DPO refresh reduces total cost by $\mathbf{10.6\times}$.}
\vspace{-0.2cm}
\label{tab:cost_main}
\footnotesize
\setlength{\tabcolsep}{6.5pt}
\renewcommand{\arraystretch}{0.7}
\begin{tabular}{@{}l c c c r@{}}
\toprule
\textbf{Variant} & \textbf{LLM Calls} & \textbf{Wall Time} & \textbf{GPU-h} & \textbf{Cost} \\
\midrule
ProfiLLM (initial)               & $\sim$1{,}460 & $\sim$6.3 hrs & 1.4 & $\$54.63$ \\
\textbf{ProfiLLM-DPO (refresh)}  & $96$          & $\sim$1.8 hrs & 1.4 & $\mathbf{\$5.13}$ \\
\bottomrule
\end{tabular}
\vspace{-0.2cm}
\end{table}

\subsection{Production Cost and Latency}
\label{subsec:cost}

\begin{figure}[t]
\centering
\includegraphics[width=0.8\linewidth]{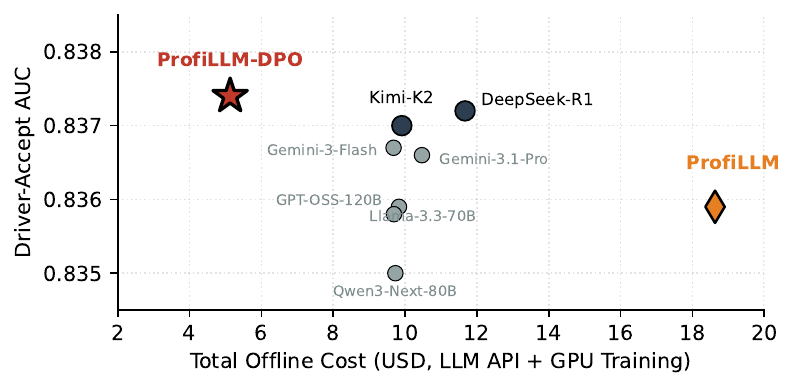}
\vspace{-0.2cm}
\caption{Cost--quality trade-off across nine LLM backbones on City~A (offline cost vs.\ driver-accept AUC).}
\Description{}
\label{fig:cost_quality_pareto}
\vspace{-0.7cm}
\end{figure}

ProfiLLM's affordability at platform scale rests on a cost structure (Table~\ref{tab:cost_main}) that shrinks at every layer. The profiling cost is first amortized by aggregation, as a single offline run yields 96 cluster profiles that cover all $348{,}464$ City~A users ($348{,}464/96\approx 3{,}630\times$ fewer profiles than per-user profiling). DPO then drives this offline cost down further, since after a one-time per-city training each routine refresh needs only $96$ single-pass LLM calls, cutting refresh cost from \$54.63 to \$5.13 ($\mathbf{10.6\times}$). At serving time the marginal cost all but vanishes, as the dispatcher runs only a deterministic cluster-rule evaluation ($<\!0.01$\,ms) and a cached embedding lookup ($<\!0.001$\,ms) per OD pair, with \emph{zero} online LLM inference, comfortably within DiDi's $200$\,ms budget. These compounding savings place ProfiLLM-DPO at the Pareto frontier of the cost--quality trade-off (Figure~\ref{fig:cost_quality_pareto}), where it dominates five of seven baseline LLMs.

\section{Related Work}
\label{sec:related}

\textbf{Ride-hailing Order Dispatching.}
Order dispatching has evolved from proximity-based greedy matching~\cite{lee2004taxi,zhang2017taxi, wang2025order, yue2024end} to RL methods capturing long-term dynamics: demand--supply forecasting with combinatorial optimization~\cite{xu2018large}, mean-field multi-agent RL~\cite{qin2020ride}, and cooperative Markov games~\cite{yang2024rethinking}. Prediction-side work models driver acceptance~\cite{wang2019adaptive} and cancellations~\cite{chen2016dynamic}. These rely on structured numerical features, leaving semantic factors unexploited; we introduce LLM-generated profiles that capture them while preserving real-time compatibility via offline--online decoupling.

\textbf{LLM Data Pipelines for Applied ML.}
Recent VLDB work operationalizes LLMs in production data systems. Closest to our setting, LEADRE~\cite{leadre2025} deploys DPO-aligned LLMs in Tencent's display-advertising pipeline, and SiriusBI~\cite{siriusbi2025} productionizes multi-round NL2SQL for enterprise BI. On the agentic side, DocETL~\cite{docetl2025}, AutoPrep~\cite{autoprep2025}, SQL-Factory~\cite{sqlfactory2025}, LLM-AutoDP~\cite{llmautodp2026}, and DBAIOps~\cite{dbaiops2026} rewrite or compose LLM-agent operators for document, tabular, SQL, and database tasks; KATS~\cite{kats2026} pairs an offline LLM knowledge graph with online retrieval. Unlike these, ProfiLLM is the first to bring an agentic LLM data pipeline into a real-time ride-hailing dispatcher (2-second cycle, sub-millisecond online budget), by confining all LLM reasoning offline and serving only cluster-level profile embeddings.

\textbf{LLM-based Agents and User Profiling.}
Tool-augmented LLM agents extend capabilities via tools (ReAct~\cite{yao2022react}, Toolformer~\cite{schick2023toolformer}, and Data-Copilot~\cite{zhang2023data}) and have been applied to traffic analysis~\cite{zhang2024trafficgpt}, though mainly for general insights rather than production prediction. We instead design a systematic tool-augmented mining workflow that yields actionable knowledge directly improving prediction.
Separately, user behavior modeling spans sequential models~\cite{kang2018self,sun2019bert4rec}, deep interest networks~\cite{zhou2018deep}, and recent LLM-based profiling~\cite{xi2024towards,liu2024once,wang2025lettingo}; in ride-hailing, prior work estimates driver value under behavioral heterogeneity~\cite{tang2021value} and models passenger demand~\cite{ke2020learning}. 
%
RLHF~\cite{ouyang2022training,bai2022training} aligns LLMs via reward modeling and PPO~\cite{schulman2017proximal} but is complex to train; DPO~\cite{rafailov2023direct} optimizes directly from preference pairs, with many extensions and task-specific variants. We contribute a novel application: constructing preference pairs from task rather than human judgment, aligning profile generation with prediction accuracy.

\section{Conclusion}
\label{sec:conclusion}

We presented ProfiLLM, a practical framework that bridges LLM-based semantic profiling with real-time ride-hailing order matching. ProfiLLM addresses three key challenges through two synergistic modules: \textit{Tool-Augmented Global Knowledge Mining}, which equips an LLM agent with 27 analytical tools to extract global knowledge, clustering rules, and regional priors; and \textit{Utility-Aligned Profile Exploration}, which iteratively refines cluster-level profiles via a lightweight prediction proxy and DPO fine-tuning. By confining all LLM inference offline and serving only pre-computed embeddings, ProfiLLM adds sub-millisecond latency. Experiments on DiDi's platform across three cities demonstrate up to +6.14\% relative AUC improvement and +4.35\% GMV gain in simulation, with a 14-day production A/B test confirming +0.47\% GMV, +0.33\% Completion Rate, and $-$0.82\% Cancel-Before-Accept rate.



\bibliographystyle{ACM-Reference-Format}
\bibliography{sample}

\clearpage
\appendix

\section{Empirical Motivation: User Behavioral Heterogeneity}
\label{app:heterogeneity}

This appendix supplements Section~\ref{sec:introduction} by quantifying \emph{how much} of order-outcome variance is governed by stable user-level traits that are invisible to structured per-order features, motivating profiling as a complementary information source. All analyses use 38 days of City~A production logs (44.3M dispatching records, 333{,}166 passengers, 12{,}128 active drivers).

\textbf{Structured features leave large unexplained behavioral variance.}
We partition grabbed orders into 100 buckets defined by (fee quintile $\times$ ETA quintile $\times$ time-of-day), so orders within a bucket share nearly identical structured features. Within these matched buckets the PCR still varies substantially across users: the average within-bucket standard deviation is 24.9\%, and the P90--P10 spread reaches 37.1 percentage points. A variance-component (ICC) analysis on the 31{,}160 passengers with $\geq 20$ orders attributes 15.8\% of PCR variance to passenger identity alone: a stable, systematic signal that per-order features cannot expose.

\begin{figure}[!h]
\centering
\includegraphics[width=\linewidth]{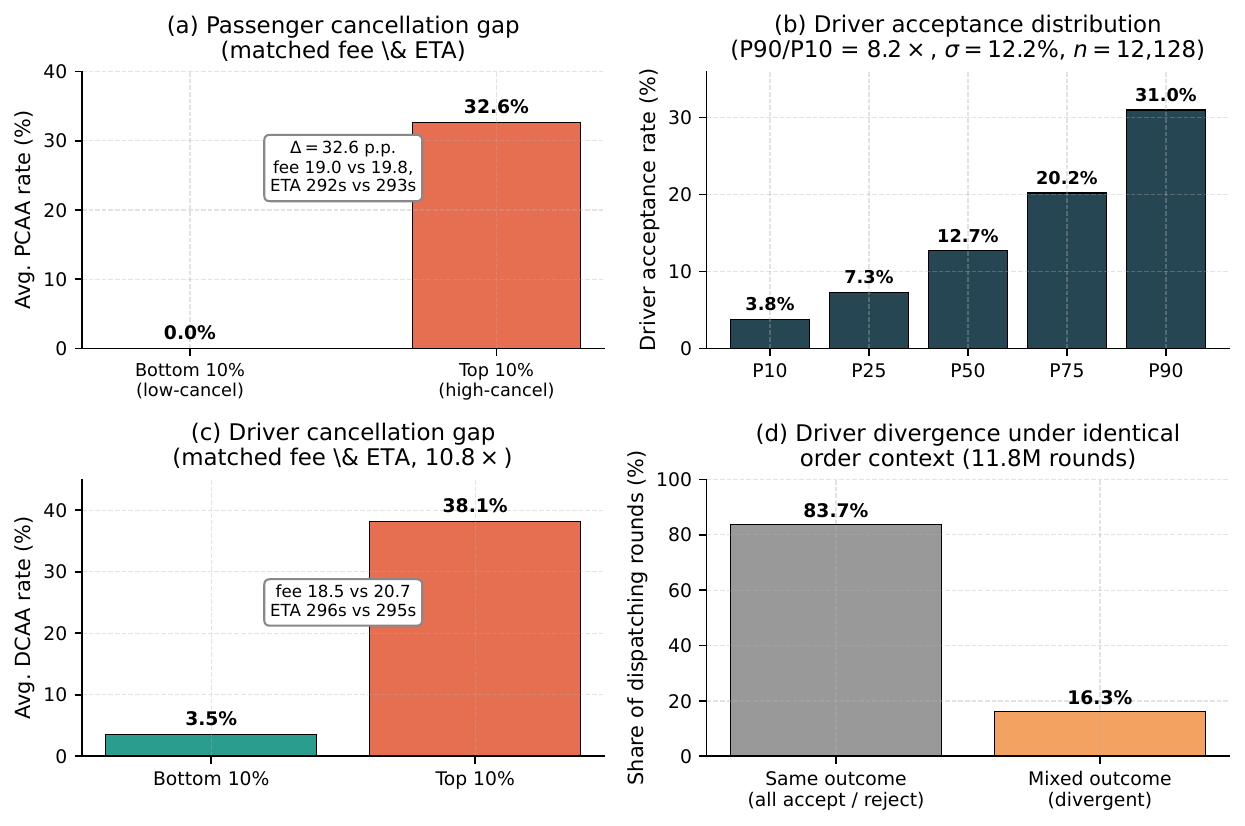}
\vspace{-0.4cm}
\caption{Behavioral heterogeneity invisible to structured features. (a) Passenger PCR decile gap under matched fee/ETA. (b) Driver DAR percentiles. (c) Driver DCR decile gap under matched conditions. (d) Share of dispatching rounds in which the same broadcasted order receives divergent accept/reject decisions from different drivers.}
\Description{A four-panel figure documenting user-level behavioral heterogeneity that structured features cannot capture. Panel (a) contrasts top- and bottom-decile passenger cancellation rates under matched fee and ETA. Panel (b) plots driver acceptance rate percentiles, showing an 8.2-times P90 over P10 spread. Panel (c) contrasts top- and bottom-decile driver cancellation rates under matched conditions. Panel (d) shows the fraction of dispatching rounds in which different drivers reach divergent accept-or-reject decisions on the same broadcasted order.}
\label{fig:user_hetero}
\vspace{-0.4cm}
\end{figure}

\textbf{Driver and passenger decile gaps under matched contexts.}
Figure~\ref{fig:user_hetero} reports complementary cuts of the data. Panel~(a) compares the most- vs.\ least-cancellation-prone passengers among those with $\geq 5$ grabbed orders: despite virtually identical average order fee (19.8 vs.\ 19.0) and ETA (293s vs.\ 292s), the former's PCR reaches 32.6\% while the latter's is 0.0\%, a 32.6\,p.p. gap. Panel~(b) plots DAR percentiles across the 12{,}128 active drivers; the P90/P10 ratio is $\mathbf{8.2\times}$ with $\sigma=12.2\%$, far beyond what order-level features can account for. Panel~(c) shows that driver cancellation behavior is similarly bimodal: top-decile DCR (38.1\%) is $\mathbf{10.8\times}$ the bottom decile (3.5\%) under matched order conditions. Panel~(d) quantifies the same effect at dispatching-round granularity: in 16.3\% of 11.8M dispatching rounds (1.93M rounds) the same broadcasted order receives mixed outcomes from different drivers (some accept, others reject), and an extreme observed case had 18 of 19 drivers reject a single order with fee~17 and ETA~384s. These patterns are stable individual traits that profiling is designed to capture.

\textbf{Heterogeneity is largely orthogonal to order-level features.}
Beyond decile-level summaries, we observe systematic individual preferences such as ETA-sensitivity heterogeneity (9.0\% of 42{,}865 multi-ETA passengers exhibit $>$30\,p.p. PCR swings across ETA bins; one passenger with 50 orders has 0\% PCR at ETA 4--6\,min but 100\% at ETA $>$8\,min) and driver price-tier selectivity (11.3\% of 10{,}687 multi-tier drivers show $\geq$20\,p.p. DAR gap between high- and low-price tiers, with a smaller 7.1\% group exhibiting the opposite preference). The pilot study in Section~\ref{sec:introduction} confirms that LLM-generated profiles convert these heterogeneity signals into 3.71\% and 9.64\% relative AUC gains on driver and passenger cancellation respectively, on top of a mature structured-feature predictor. Together, these analyses establish that user-side variance is large, systematic, and complementary to per-order features, precisely the gap ProfiLLM is designed to close.

\textbf{Long-tail prevalence amplifies the need for clustering.}
Of the 333{,}166 passengers in the analysis window, 44.9\% appear in $\leq 3$ orders and 59.3\% in $\leq 5$ orders (consistent with Figure~\ref{fig:longtail}). For these users no individual-level behavioral signal can be reliably estimated, motivating ProfiLLM's adaptive cluster-level profiling that transfers knowledge from data-rich groups to data-sparse individuals.

\begin{table*}[t]
\centering
\caption{Categorization of analytical tools in ProfiLLM.}
\label{tab:tools}
\small
\renewcommand{\arraystretch}{1.1}
\resizebox{\textwidth}{!}{%
\begin{tabular}{>{\centering\arraybackslash}m{1.5cm}clp{4.8cm}|>{\centering\arraybackslash}m{1.5cm}clp{4.8cm}}
\toprule
\textbf{Category} & \textbf{\#} & \textbf{Tool} & \textbf{Description} & \textbf{Category} & \textbf{\#} & \textbf{Tool} & \textbf{Description} \\
\midrule
\multirow{4}{*}{Statistical} & \multirow{4}{*}{4} & \texttt{AggregateStats} & Calculate aggregate statistics by dimension & \multirow{6}{*}{\shortstack{Spatio-\\temporal}} & \multirow{6}{*}{6} & \texttt{DetectPeakPeriods} & Detect peak periods for metrics \\
 & & \texttt{CompareSegments} & Compare segments on specific metrics & & & \texttt{DayOfWeekPattern} & Analyze weekday vs weekend patterns \\
 & & \texttt{UserClustering} & K-Means clustering on user behavior & & & \texttt{HourlyPattern} & Analyze 24-hour detailed patterns \\
 & & \texttt{FeatureImportance} & Analyze key features for target metric & & & \texttt{SpatialHotspot} & Identify spatial distribution hotspots \\
\cline{1-4}
\multirow{5}{*}{Causal} & \multirow{5}{*}{5} & \texttt{CausalDiscovery} & Discover causal relationships & & & \texttt{ODFlowAnalysis} & Analyze origin-destination flow patterns \\
 & & \texttt{CounterfactualAnalysis} & Perform ``what if'' analysis & & & \texttt{RegionCharacteristics} & Analyze regional characteristic profiles \\
\cline{5-8}
 & & \texttt{ChainOfMining} & Multi-stage iterative analysis & \multirow{7}{*}{Contextual} & \multirow{7}{*}{7} & \texttt{SupplyDemandAnalysis} & Analyze supply-demand balance \\
 & & \texttt{UncertaintyAwareMining} & Provide confidence intervals & & & \texttt{WaitTimeFactors} & Analyze factors affecting wait time \\
 & & \texttt{ContrastiveAnalysis} & Compare similar groups for differences & & & \texttt{MatchingEfficiency} & Analyze order matching efficiency \\
\cline{1-4}
\multirow{3}{*}{Knowledge} & \multirow{3}{*}{3} & \texttt{GlobalCausalRules} & Discover global causal rules & & & \texttt{DetectAnomalies} & Detect anomalies in metrics \\
 & & \texttt{GlobalBenchmarks} & Generate benchmarks for classification & & & \texttt{SpecialPeriodAnalysis} & Analyze holiday/event patterns \\
 & & \texttt{ProfileKnowledgeBase} & Generate profile usage guide & & & \texttt{WeatherFactorAnalysis} & Analyze weather impact on metrics \\
\cline{1-4}
\multirow{2}{*}{Validation} & \multirow{2}{*}{2} & \texttt{ValidationDiscovery} & Validate discovered patterns & & & \texttt{WeatherScenario} & Analyze specific weather scenarios \\
 & & \texttt{ConclusionValidation} & Cross-validate conclusions & & & & \\
\bottomrule
\end{tabular}%
}
\end{table*}

\section{Representative Case Studies}
\label{app:case_study}

To illustrate the kind of behavioral heterogeneity that ProfiLLM's profile embeddings are designed to capture, we walk through two representative cases drawn from City~A production logs.

Driver case: divergent acceptance under nearly identical order context. In a single dispatching round, order \texttt{2209******7356} (fee \$17.0, ETA 384\,s, evening hour~17, origin/destination both in the city center) was broadcast to 19 drivers. Only one driver accepted; the other 18 rejected. The accepting driver had a baseline DAR of $0.089$ and historical completion rate of $0.289$; a representative rejecting driver had DAR $0.114$ and completion rate $0.222$, so a predictor relying purely on order-side and aggregate driver-side structured statistics would have ranked the rejecting driver higher. Across the full log, 16.3\% of dispatching rounds (1.93\,M of 11.8\,M batches) exhibit such mixed-outcome broadcasts where the same order receives divergent accept/reject decisions from different drivers under near-identical order-side context (Appendix~\ref{app:heterogeneity}). ProfiLLM's cluster-level driver profile encodes precisely this kind of identity-driven pattern, e.g., the accepting driver's cluster is characterized by a willingness to take evening city-center orders with mid-tier fares, while the rejecting drivers' clusters are not.

Passenger case: ETA-sensitivity stratification under matched fare. The most cancellation-prone passengers in City~A (18{,}394 passengers, average PCR $32.6\%$) book at average fare \$19.8 and ETA 293\,s; a comparison group of 97{,}579 passengers exhibits PCR $0.0\%$ under almost identical conditions (fare \$19.0, ETA 292\,s). A finer-grained example: passenger A (50 grabbed orders) cancels 0\% of orders with ETA in 4--6\,min but 100\% of orders with ETA $>$\,8\,min, while the platform average rises only from $6.9\%$ to $11.3\%$ across the same ETA band. This per-passenger ETA-tolerance threshold is invisible to structured features that report only the absolute ETA value; ProfiLLM's cluster-level passenger profile materializes such latent tolerance patterns as part of the cluster's PROFILE narrative (\eg, commute-hour passengers who tolerate $\leq$\,6\,min wait but defect at longer ETAs), feeding the prediction model a discriminative signal the structured-feature baseline cannot express.

\section{Background}
\label{subsec:background}

\noindent\textbf{Tool-Augmented LLM Agents.}
Recent advances have demonstrated that LLMs can effectively leverage external tools to accomplish complex tasks beyond their inherent capabilities~\cite{yao2022react,ning2025dima}. 
A tool-augmented LLM agent operates by iteratively generating reasoning traces and invoking tools based on intermediate observations. 
Formally, given an input query $q$ and a tool set $\mathcal{T} = \{t_1, \ldots, t_{|\mathcal{T}|}\}$, the agent produces a trajectory $\tau = \{(a_i, r_i)\}_{i=1}^{L}$, where $a_i$ denotes an action (either reasoning or tool invocation) and $r_i$ denotes the corresponding observation or tool result. 
This paradigm enables LLMs to analyze data at scales far exceeding their context window limitations.

\textbf{Direct Preference Optimization (DPO).}
DPO~\cite{rafailov2023direct} provides an efficient approach to align LLM outputs with human preferences without explicit reward modeling. Given preference pairs $(x, y_w, y_l)$ where $y_w$ is preferred over $y_l$ for input $x$, DPO directly optimizes the policy $\pi_\theta$ via:
\begin{equation}
    \mathcal{L}_{\text{DPO}} = -\mathbb{E}_{(x, y_w, y_l)} \left[ \log \sigma \left( \beta \log \frac{\pi_\theta(y_w|x)}{\pi_{\text{ref}}(y_w|x)} - \beta \log \frac{\pi_\theta(y_l|x)}{\pi_{\text{ref}}(y_l|x)} \right) \right]
\label{eq:dpo}
\end{equation}
where $\pi_{\text{ref}}$ is a reference policy (typically the supervised fine-tuned model), $\beta$ controls the deviation from the reference, and $\sigma(\cdot)$ is the sigmoid function. In our context, we extend DPO to align profile generation with downstream prediction utility.

\begin{algorithm}[t]
\caption{Tool-Augmented Global Knowledge Mining}
\label{alg:mining}
\small
\begin{algorithmic}[1]
\Require Historical data $\mathcal{H}$, Tool set $\mathcal{T}$, LLM agent $\mathcal{M}$
\Ensure Global knowledge $\mathcal{K}$, Clustering rules $\mathcal{A}$, Regional priors $\mathcal{R}$
\State \textcolor{gray}{\textit{// Phase 1: Explore}}
\State $\textit{findings}_1 \gets \emptyset$
\For{$t \in \mathcal{T}_{\textit{basic}}$}
    \State $\textit{result} \gets t.\text{execute}(\mathcal{H})$
    \State $\textit{findings}_1 \gets \textit{findings}_1 \cup \mathcal{M}.\text{interpret}(\textit{result})$
\EndFor
\State \textcolor{gray}{\textit{// Phase 2: Deepen}}
\State $\textit{directions} \gets \mathcal{M}.\text{identify\_directions}(\textit{findings}_1)$
\State $\textit{findings}_2 \gets \emptyset$
\For{$\textit{dir} \in \textit{directions}$}
    \State $\textit{tools} \gets \mathcal{M}.\text{select\_tools}(\textit{dir}, \mathcal{T})$
    \State $\textit{result} \gets \text{ExecuteToolChain}(\textit{tools}, \mathcal{H})$
    \State $\textit{findings}_2 \gets \textit{findings}_2 \cup \mathcal{M}.\text{analyze}(\textit{result})$
\EndFor
\State \textcolor{gray}{\textit{// Phase 3: Validate}}
\State $\textit{candidates} \gets \mathcal{M}.\text{extract\_hypotheses}(\textit{findings}_1 \cup \textit{findings}_2)$
\State $\textit{validated} \gets \emptyset$
\For{$\textit{hyp} \in \textit{candidates}$}
    \State $\textit{pval}, \textit{eff} \gets \texttt{validate\_hypothesis}(\textit{hyp}, \mathcal{H})$
    \If{$\textit{pval} < \alpha$ \textbf{and} $|\textit{eff}| > \epsilon$}
        \State $\textit{validated} \gets \textit{validated} \cup \{(\textit{hyp}, \textit{eff})\}$
    \EndIf
\EndFor
\State \textcolor{gray}{\textit{// Phase 4: Synthesize}}
\State $\mathcal{K} \gets \mathcal{M}.\text{synthesize\_knowledge}(\textit{validated})$
\State $\mathcal{A} \gets \mathcal{M}.\text{generate\_clustering\_rules}(\textit{findings}_2)$
\State $\mathcal{R} \gets \texttt{compute\_regional\_priors}(\mathcal{H}, \mathcal{G})$
\State \Return $\mathcal{K}, \mathcal{A}, \mathcal{R}$
\end{algorithmic}
\end{algorithm}

\section{Analytical Tool Details}
\label{app:tools}

Table~\ref{tab:tools} presents the complete categorization of the 27 analytical tools used in the Tool-Augmented Global Knowledge Mining module. These tools are organized into six categories based on their analytical functionality: (1) \textbf{Statistical} tools for computing aggregate statistics, comparing segments, clustering users, and identifying important features; (2) \textbf{Causal} tools for discovering causal relationships, performing counterfactual analysis, and conducting multi-stage iterative mining with uncertainty quantification; (3) \textbf{Knowledge} tools for extracting global causal rules, generating classification benchmarks, and constructing profile knowledge bases; (4) \textbf{Validation} tools for verifying discovered patterns and cross-validating conclusions; (5) \textbf{Spatiotemporal} tools for detecting peak periods, analyzing day-of-week and hourly patterns, identifying spatial hotspots, and examining origin-destination flows; and (6) \textbf{Contextual} tools for analyzing supply-demand balance, wait time factors, matching efficiency, anomalies, special periods, and weather impacts. All tools are designed to accept structured parameters and return interpretable results that the LLM agent can reason over, enabling composable tool chains for complex analytical queries.

\section{Algorithm Pseudocode}
\label{app:algorithms}

This section provides the detailed pseudocode for the two core procedures in ProfiLLM.

\textbf{Algorithm~\ref{alg:mining}} formalizes the Tool-Augmented Global Knowledge Mining workflow described in Section~\ref{subsec:knowledge_mining}. The procedure follows the four-phase Explore-Deepen-Validate-Synthesize paradigm. In the Explore phase, the agent invokes basic statistical tools to obtain an initial understanding of the data landscape. The Deepen phase identifies promising analytical directions from preliminary findings and applies targeted tool chains for focused investigation. The Validate phase subjects each discovered pattern to statistical hypothesis testing, retaining only findings with $p$-value below threshold $\alpha$ and effect size exceeding $\epsilon$. Finally, the Synthesize phase consolidates validated findings into three structured outputs: global knowledge $\mathcal{K}$, user clustering rules $\mathcal{A}$, and regional supply-demand priors $\mathcal{R}$.

\textbf{Algorithm~\ref{alg:profile}} details the DPO-Aligned Profile Exploration procedure described in Section~\ref{subsec:dpo_profile}. Given a user cluster $a$ with its aggregated history and the global knowledge base, the algorithm first generates $K$ diverse candidate profiles and evaluates each via the LOGIC-rule-based utility proxy (Eq.~\ref{eq:utility_gain}). The best-performing candidate then undergoes iterative refinement for $T$ rounds: at each iteration, prediction errors are analyzed to produce targeted feedback, and the LLM generates an improved profile conditioned on this feedback. Throughout the process, all candidate profiles are compared pairwise to construct preference pairs with a threshold $\gamma$, which are subsequently used for DPO fine-tuning to align the LLM's generation capability with downstream prediction utility.

\begin{algorithm}[t]
\caption{Utility-Aligned Profile Exploration}
\label{alg:profile}
\small
\begin{algorithmic}[1]
\Require Cluster $a$, Aggregated history $\mathcal{H}_a$, Global knowledge $\mathcal{K}$, LLM $\mathcal{M}$
\Ensure Optimal profile $\textit{profile}_a^*$, Preference pairs $\mathcal{P}_a$
\State \textcolor{gray}{\textit{// Initial candidate generation}}
\State $\{\textit{profile}_a^{(k)}\}_{k=1}^K \gets \mathcal{M}.\text{generate}(\mathcal{H}_a, \mathcal{K}, K)$
\State \textcolor{gray}{\textit{// Evaluate initial candidates via LOGIC rules}}
\For{$k = 1$ \textbf{to} $K$}
    \State $\textit{LOGIC}_a^{(k)} \gets \text{extract\_logic}(\textit{profile}_a^{(k)})$
    \State $\Delta_a^{(k)} \gets \text{EvaluateUtility}(\textit{LOGIC}_a^{(k)}, \mathcal{H}_a)$ \Comment{Eq.~\ref{eq:utility_gain}}
\EndFor
\State $k^* \gets \arg\max_k \Delta_a^{(k)}$
\State $\textit{profile}_a^{\textit{best}} \gets \textit{profile}_a^{(k^*)}$; $\Delta_a^{\textit{best}} \gets \Delta_a^{(k^*)}$
\State \textcolor{gray}{\textit{// Iterative refinement}}
\For{$t = 1$ \textbf{to} $T$}
    \State $\textit{feedback} \gets \text{AnalyzeErrors}(\textit{LOGIC}_a^{\textit{best}}, \mathcal{H}_a)$
    \State $\{\textit{profile}_a^{(t,k)}\}_{k=1}^{K} \gets \mathcal{M}.\text{refine}(\textit{profile}_a^{\textit{best}}, \textit{feedback}, K)$
    \For{$k = 1$ \textbf{to} $K$}
        \State $\Delta_a^{(t,k)} \gets \text{EvaluateUtility}(\textit{profile}_a^{(t,k)}, \mathcal{H}_a)$
    \EndFor
    \State $k^{\dagger} \gets \arg\max_k \Delta_a^{(t,k)}$
    \If{$\Delta_a^{(t,k^{\dagger})} > \Delta_a^{\textit{best}}$}
        \State $\textit{profile}_a^{\textit{best}} \gets \textit{profile}_a^{(t,k^{\dagger})}$; $\Delta_a^{\textit{best}} \gets \Delta_a^{(t,k^{\dagger})}$
    \EndIf
\EndFor
\State $\textit{profile}_a^* \gets \textit{profile}_a^{\textit{best}}$
\State \textcolor{gray}{\textit{// Construct preference pairs for DPO}}
\State $\mathcal{P}_a \gets \{(\mathcal{H}_a, \textit{profile}_w, \textit{profile}_l) : \Delta_w > \Delta_l + \gamma\}$
\State \Return $\textit{profile}_a^*$, $\mathcal{P}_a$
\end{algorithmic}
\end{algorithm}

\section{Discovered Cluster Archetypes}
\label{app:archetypes}

ProfiLLM's clustering rules are \emph{not} manually specified. They are produced by the LLM agent during the Synthesize phase of Tool-Augmented Global Knowledge Mining (Algorithm~\ref{alg:mining}), which interprets cluster centroids and their $z$-score deviations from the population mean to assign meaningful archetype labels. Drivers and passengers are clustered separately into $A=A_D\cup A_P$, where the agent settled on $|A_D|=28$ driver clusters and $|A_P|=49$ passenger clusters in our main experiments. Clustering uses dozens of behavioral features, including order volume and frequency, acceptance/grab rate, cancellation rate and patterns, completion rate, average fee and price sensitivity, active-time distribution, trip distance and duration, spatial activity patterns, and derived ratios (e.g., cancel-to-complete ratio, peak-hour share).

\begin{figure}[t]
\centering
\includegraphics[width=\linewidth]{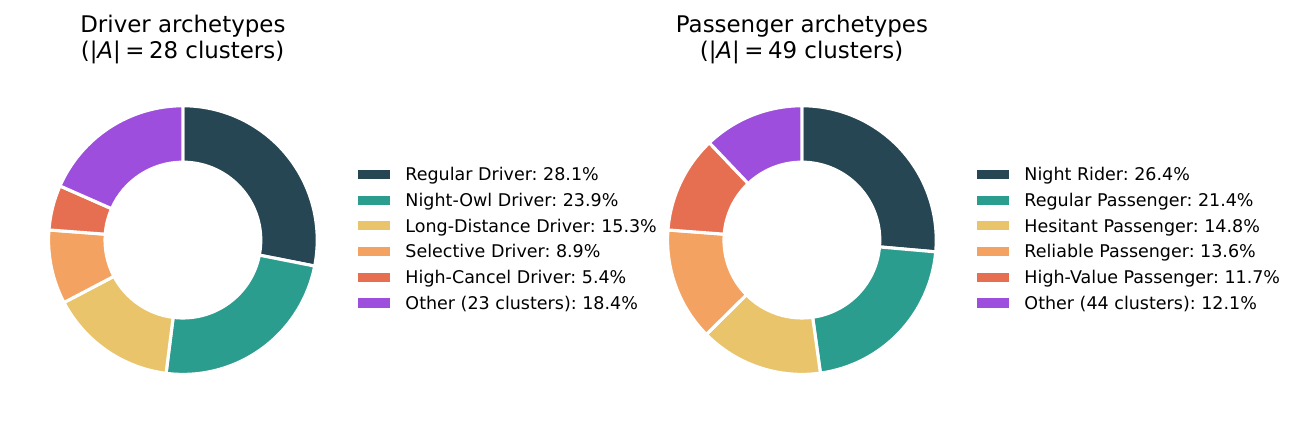}
\vspace{-0.4cm}
\caption{Population share of the top-5 LLM-discovered archetypes in City~A for drivers ($|A_D|=28$) and passengers ($|A_P|=49$), with the remaining clusters aggregated for clarity.}
\Description{Two bar charts showing the population share of the top-5 LLM-discovered archetypes for drivers and passengers in City~A, with all remaining clusters aggregated into an "Other" bar. The dominant driver archetypes (Regular, Night-Owl, Long-Distance, Selective, High-Cancellation) together cover 81.6\% of drivers, and the top-5 passenger archetypes cover 87.9\% of passengers.}
\label{fig:archetypes}
\end{figure}

Figure~\ref{fig:archetypes} reports the population share of the top-5 archetypes for each role, and Table~\ref{tab:archetypes} lists their defining characteristics. The dominant driver groups (Regular, Night-Owl, Long-Distance, Selective, High-Cancel) together cover 81.6\% of the active driver fleet, while the corresponding top-5 passenger groups cover 87.9\% of the passenger population. Tail clusters retain meaningful distinctions (e.g., airport-specialist drivers, time-sensitive commuter passengers) that the LLM agent labels using domain-grounded heuristics from the mined global knowledge.

\begin{table}[t]
\centering
\caption{Representative archetypes discovered by the LLM agent in City~A. Listed signatures are cluster-centroid characteristic values (salient deviations from the population mean) used by the agent for labeling, not realized post-acceptance PCR/DCR rates.}
\label{tab:archetypes}
\footnotesize
\setlength{\tabcolsep}{3.5pt}
\renewcommand{\arraystretch}{1.05}
\resizebox{\linewidth}{!}{%
\begin{tabular}{@{}l l l@{}}
\toprule
\textbf{Archetype} & \textbf{Share} & \textbf{Key behavioral signatures} \\
\midrule
\multicolumn{3}{l}{\textit{Driver}} \\
Regular Driver         & 28.1\% & Moderate activity, morning-dominant hours \\
Night-Owl Driver       & 23.9\% & Avg.\ hour 16.1, high evening concentration \\
Long-Distance Driver   & 15.3\% & Avg.\ fee 20.8, avg.\ trip 7.7\,km \\
Selective Driver       &  8.9\% & Low volume but 44.6\% grab rate \\
High-Cancellation Driver &5.4\% & Cancel rate 84.5\%, low completion \\
\midrule
\multicolumn{3}{l}{\textit{Passenger}} \\
Night Rider            & 26.4\% & Avg.\ hour 16.8, evening-dominant \\
Regular Passenger      & 21.4\% & Moderate volume, morning-focused \\
Hesitant Passenger     & 14.8\% & Cancel rate 69.6\%, low reliability \\
Reliable Passenger     & 13.6\% & Low volume, completion rate 46.9\% \\
High-Value Passenger   & 11.7\% & Avg.\ fare 34.0, avg.\ trip 14.1\,km \\
\bottomrule
\end{tabular}}
\end{table}

\begin{figure*}[t]
  \centering
  \begin{minipage}[t]{0.48\textwidth}
    \centering
    \includegraphics[width=\linewidth]{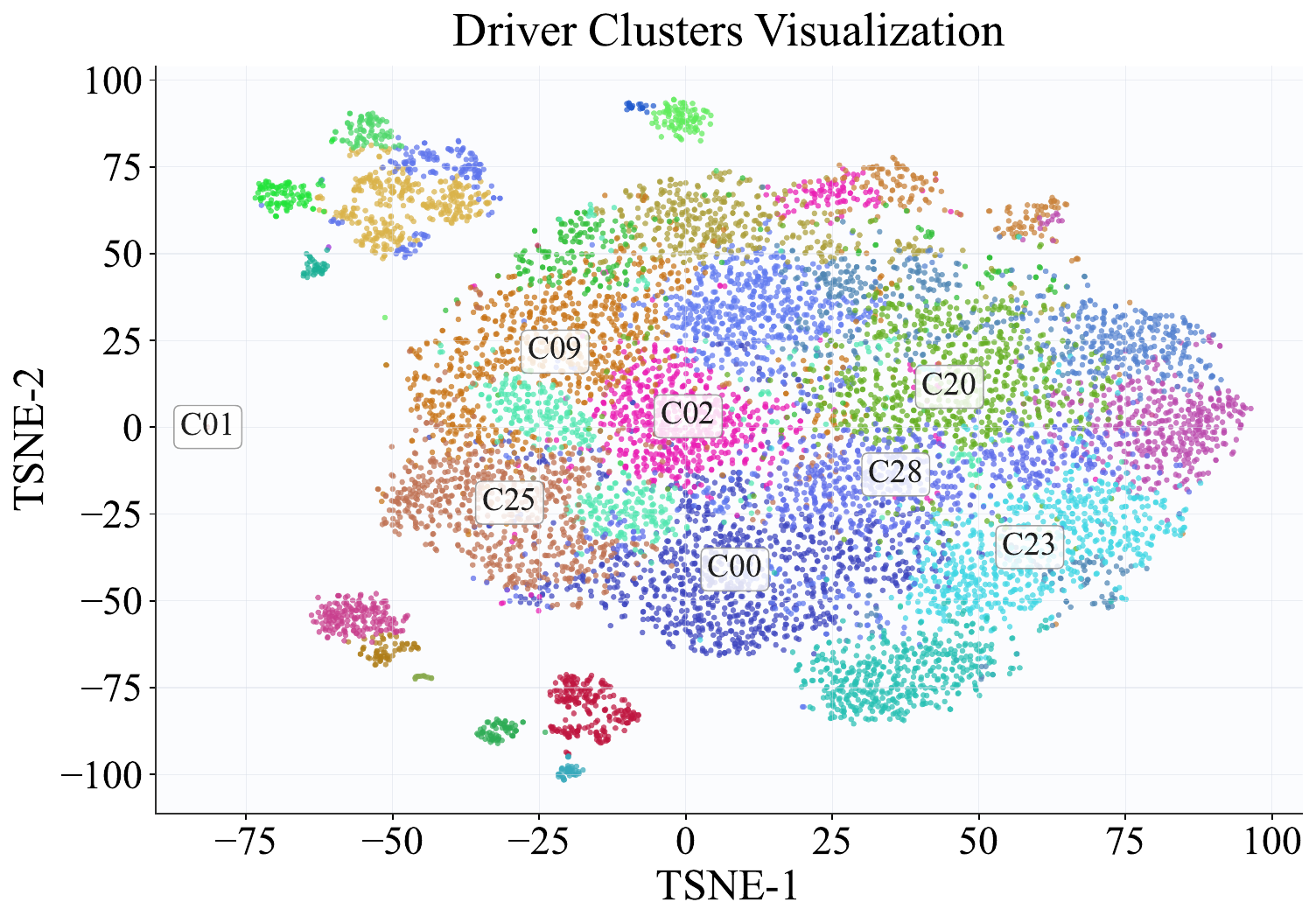}
    \vspace{4pt}\centerline{\small (a) Driver clusters}
    \label{fig:cluster_vis_driver}
  \end{minipage}
  \hfill
  \begin{minipage}[t]{0.48\textwidth}
    \centering
    \includegraphics[width=\linewidth]{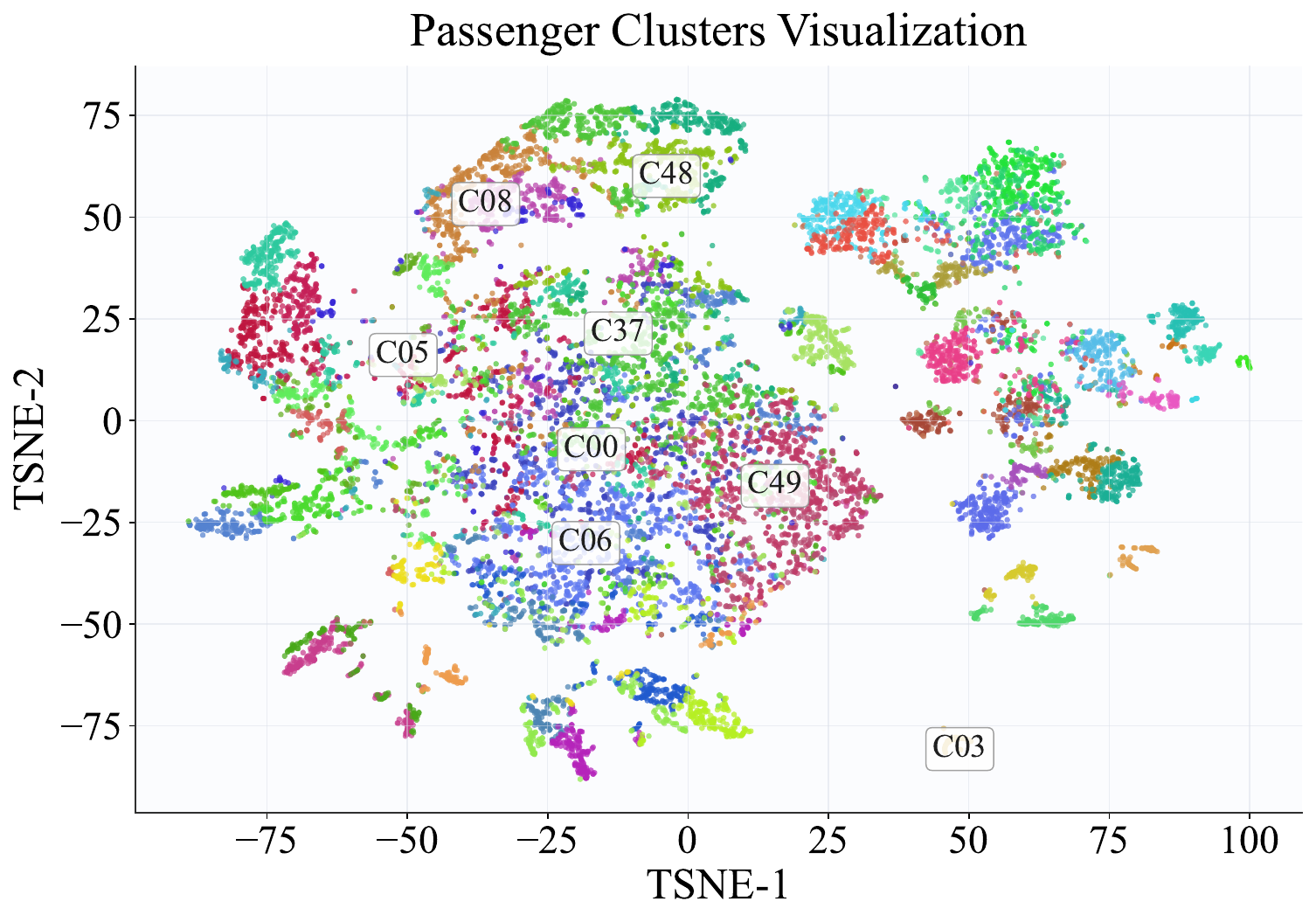}
    \vspace{4pt}\centerline{\small (b) Passenger clusters}
    \label{fig:cluster_vis_pax}
  \end{minipage}
  \vspace{-0.6cm}
  \caption{t-SNE visualization of user cluster embeddings in City~A, showing (a) driver clusters and (b) passenger clusters. Each point represents a user colored by cluster assignment; the top-8 largest clusters are labeled at their median positions.}
  \Description{Two side-by-side 2D t-SNE scatter plots of user behavioral embeddings in City~A, with points colored by cluster assignment and the top-8 largest clusters labeled at their median positions. The driver plot (a) shows compact, well-separated clusters, while the passenger plot (b) shows broader, more overlapping clusters indicating greater behavioral diversity.}
  \label{fig:cluster_vis}
\end{figure*}

\subsection{User Cluster Embedding Visualization}
\label{app:cluster_vis}
To qualitatively assess whether the clustering rules $\mathcal{A}$ yield behaviorally separable groups, we visualize user-level embedding distributions via t-SNE~\cite{maaten2008visualizing}.
For each user $u \in \mathcal{P} \cup \mathcal{D}$, we extract a behavioral feature vector from $\mathcal{H}_u$ and assign them to cluster $a^*(u)$ via rules~$\mathcal{A}$.
Before dimensionality reduction, we apply core filtering (retaining the nearest 80\% of points to each cluster centroid) and stratified sampling.
The filtered embeddings are standardized, reduced to 50 dimensions via PCA, and projected to 2D with t-SNE (perplexity\,=\,35, PCA initialization, seed\,=\,42).

Figure~\ref{fig:cluster_vis} presents the resulting scatter plots for City~A, with the top-8 largest clusters annotated at their median positions.
The driver embedding space (Figure~\ref{fig:cluster_vis}~(a)) exhibits well-separated clusters corresponding to discovered archetypes, including \textit{Regular Drivers}~(C00), \textit{Night Owls}~(C01), and \textit{Long-Distance Drivers}~(C02).
The passenger embedding space (Figure~\ref{fig:cluster_vis}~(b)) similarly reveals distinct groups such as \textit{Night Riders}, \textit{Regular Passengers}, and \textit{Hesitant Passengers} with elevated cancellation tendencies, though with more overlap reflecting higher behavioral diversity on the passenger side.
The clear visual separation confirms that the LLM-agent-derived clustering rules partition users into behaviorally coherent groups, supporting cluster-level profiles as effective proxies for individual user behavior in downstream outcome prediction.

\section{Dispatching Simulator Architecture}
\label{app:simulator}

We summarize the design choices that make our simulator a faithful and reproducible offline counterpart to production dispatcher.

\textbf{Replay-based discrete-event environment.}
The simulator replays five full days of historical order arrivals and driver availability per city, rather than generating synthetic demand. All orders arrive at their actual timestamps with real origin, destination, dynamic pricing (including surge multipliers), and over 30 contextual features. Driver positions are initialized from actual GPS trajectory logs, and each driver's online-duration distribution is reconstructed from historical records. The geographic space uses the same production grid system, preserving real-world spatiotemporal distributions of demand, supply, and traffic.

\textbf{Production-identical dispatching cadence and matching.}
The simulator operates in 2-second dispatching cycles. At each cycle, candidate OD pairs are enumerated within a 1{,}500-meter pickup radius; the production STR (Spatio-Temporal Revenue) formula scores each pair using Accept/P-Cancel/D-Cancel predictions, pickup cost, and order value with production-calibrated weights; and the optimal bipartite assignment is solved via a Lagrangian-relaxation variant of Kuhn--Munkres, identical to the algorithm deployed in production.

\textbf{Production routing API integration.}
The simulator queries DiDi's production routing service via Thrift RPC to obtain live-traffic-aware ETA and pickup distance (with 25th/75th-percentile confidence intervals), eliminating a major source of bias that approximate offline routing would introduce.

\textbf{Three-stage stochastic outcome simulation.}
Rather than sampling a single completion probability, the simulator implements a three-stage sequential decision process: (i)~Accept sampling (driver acceptance); (ii)~conditional D-Cancel sampling; (iii)~P-Cancel sampling. An order completes \emph{only if} the driver accepts and neither party cancels. Failed orders re-enter the pending queue for re-dispatch up to a 180-second patience timeout, producing realistic re-dispatch cascades. This three-stage design captures the feedback loop in which a better predictor yields better matches, fewer rejections, fewer re-dispatches, and higher system-wide efficiency.

\textbf{Dynamic state evolution.}
Drivers transition between \emph{idle}, \emph{en-route-to-pickup}, and \emph{in-service} states with real-time features refreshed every cycle, ensuring that downstream feature distributions remain consistent with production. Orders follow realistic lifecycle transitions with timeouts, and grid-level supply-demand statistics support spatial-aware matching weights.

\textbf{Validation against production.}
Comparing simulation (Table~\ref{tab:results}, City~A) with the 14-day online A/B test (Figure~\ref{fig:online_ab}) shows directional consistency on every core metric (GMV, CR, PCR, DCR all moving in the desired direction), though the online deltas are several times smaller (\eg \textbf{+0.47\%} vs.\ \textbf{+4.02\%} GMV, a $\sim$8--9$\times$ gap). Simulation magnitudes exceed online deltas because the simulator both scores candidate matches and samples their outcomes from the \emph{same} model, grading each policy on its own beliefs in a closed loop that omits the live confounders (driver multi-homing, exogenous demand shocks, concurrent experiments) which attenuate real treatment effects; this downward bias is well documented for two-sided-marketplace experiments~\cite{holtz2020limiting}. This direction-preserving, magnitude-inflating behavior is characteristic of replay-based ride-hailing simulators~\cite{xu2018large,qin2020ride,yang2024rethinking} and mirrors the broader offline-to-online gap, in which offline metrics overstate online performance while better preserving method ranking~\cite{hidasi2023widespread}. The key validation criterion is therefore directional consistency and method ranking preservation, both of which our simulator delivers.

\begin{figure}[t]
    \centering
    \includegraphics[width=\linewidth]{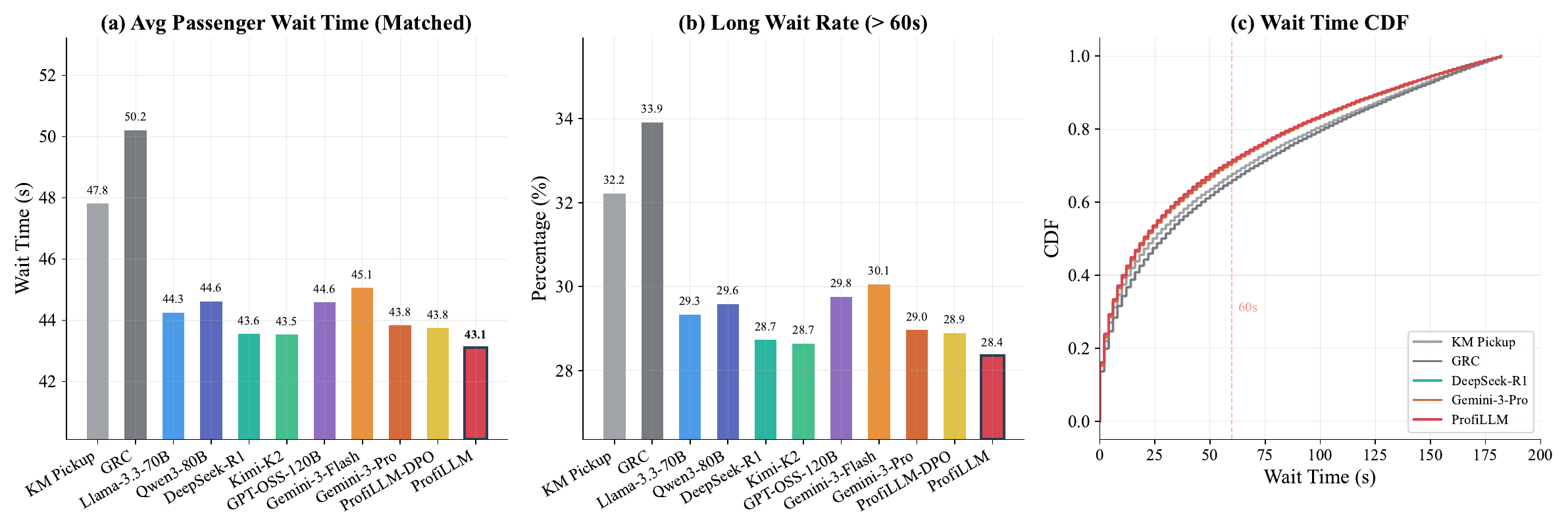}
    \caption{Passenger wait time analysis in the dispatching simulator. (a) Average wait time for matched orders. (b) Proportion of orders with wait time exceeding 60 seconds. (c) Cumulative distribution function (CDF) of wait times for representative strategies.}
    \Description{A three-panel figure comparing passenger wait times across dispatching strategies. Panels (a) and (b) are bar charts of average wait time and the share of orders exceeding 60 seconds. Panel (c) is a CDF plot in which the ProfiLLM curve sits to the left of all baselines across the entire distribution, indicating shorter wait times at every quantile.}
    \label{fig:passenger_wait_time}
\end{figure}

\section{Passenger Wait Time Analysis}
\label{sec:appendix_wait_time}

We further analyze passenger wait time in the dispatching simulator to evaluate user experience beyond platform-level metrics. As shown in Figure~\ref{fig:passenger_wait_time}, the two non-LLM baselines exhibit the highest average wait times (GRC: 50.2s, KM Pickup: 47.8s), while all LLM-based strategies achieve notably lower values ranging from 43.1s to 45.1s. ProfiLLM attains the lowest average wait time of 43.1s, representing reductions of 9.8\% over KM Pickup and 14.1\% over GRC. The long-wait rate ($>$60s) follows a consistent pattern: GRC and KM Pickup reach 33.9\% and 32.2\%, respectively, whereas ProfiLLM reduces this to 28.4\%. The CDF curves further confirm a systematic leftward shift for ProfiLLM across all quantiles, indicating that the improvement is not driven by a small subset of orders but reflects enhanced dispatching efficiency throughout. These results complement the GMV and CR metrics in Section~\ref{subsec:overall}, demonstrating that utility-aligned user profiles improve not only platform revenue but also passenger-perceived service quality.

\section{Hourly Performance Analysis}
\label{sec:appendix_hourly}

\begin{figure}[t]
    \centering
    \includegraphics[width=\linewidth]{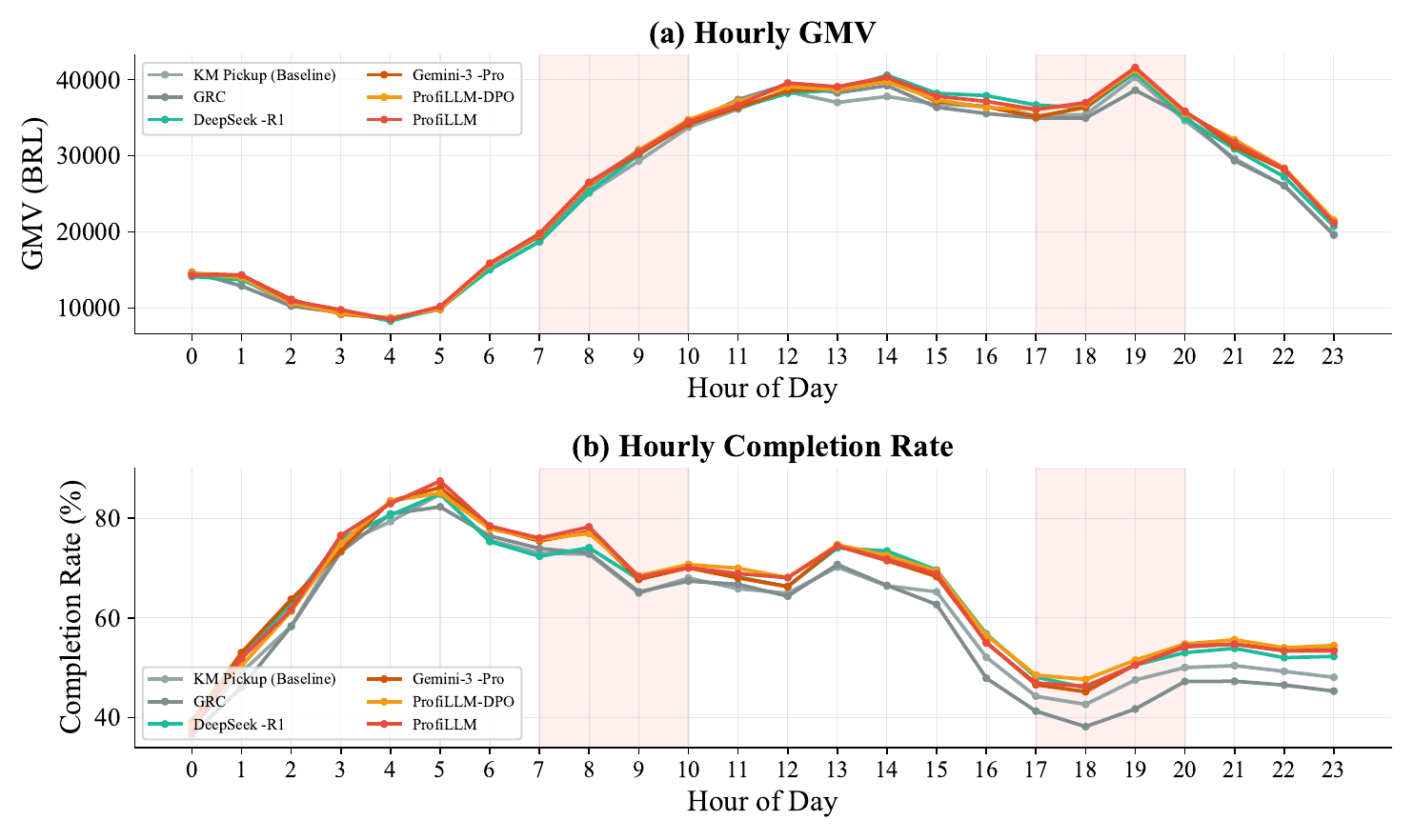}
    \vspace{-0.4cm}
    \caption{Hourly GMV and CR in the dispatching simulator. Shaded regions indicate peak hours (morning 7--10, noon 11--14, evening 17--20).}
    \Description{Two stacked line plots showing hourly GMV and completion rate (CR) across a 24-hour day in the dispatching simulator, with morning, noon, and evening peak windows shaded. ProfiLLM and ProfiLLM-DPO curves stay above the baselines throughout the day and pull further ahead during the evening peak, while GRC's CR drops sharply during evening hours.}
    \label{fig:hourly_performance}
\end{figure}

To understand how different strategies perform across varying demand conditions, we analyze hourly GMV and CR in the dispatching simulator. As shown in Figure~\ref{fig:hourly_performance}, all strategies follow the same demand pattern with peaks around noon (12--13h) and evening (18--19h), and a trough in the early morning (4--5h). ProfiLLM and ProfiLLM-DPO consistently outperform baselines throughout the day, with the gap most pronounced during evening peak hours (17--20h) when supply-demand imbalance intensifies and accurate outcome prediction becomes critical. Notably, GRC suffers a sharp CR drop during evening hours (falling below 40\%), likely due to its cooperative game formulation struggling under severe supply constraints, whereas ProfiLLM maintains stable performance above 50\%. The consistent hourly advantage confirms that utility-aligned user profiles provide robust improvements across diverse operational conditions rather than benefiting only specific time periods.

\section{Cluster-Count Sensitivity}
\label{app:cluster_sensitivity}

To verify that the framework is robust to the granularity of clustering, we evaluate 11 cluster configurations on City~A by varying one side while fixing the other: $|A_D|\in\{8,16,32,64,128,256\}$ at $|A_P|=64$, and $|A_P|\in\{8,16,32,64,128,256\}$ at $|A_D|=32$. Each configuration is evaluated against 9 LLM backbones (the 7 baselines in Table~\ref{tab:auc_results} plus ProfiLLM and ProfiLLM-DPO), totaling 99 runs.

\begin{figure}[t]
\centering
\includegraphics[width=\linewidth]{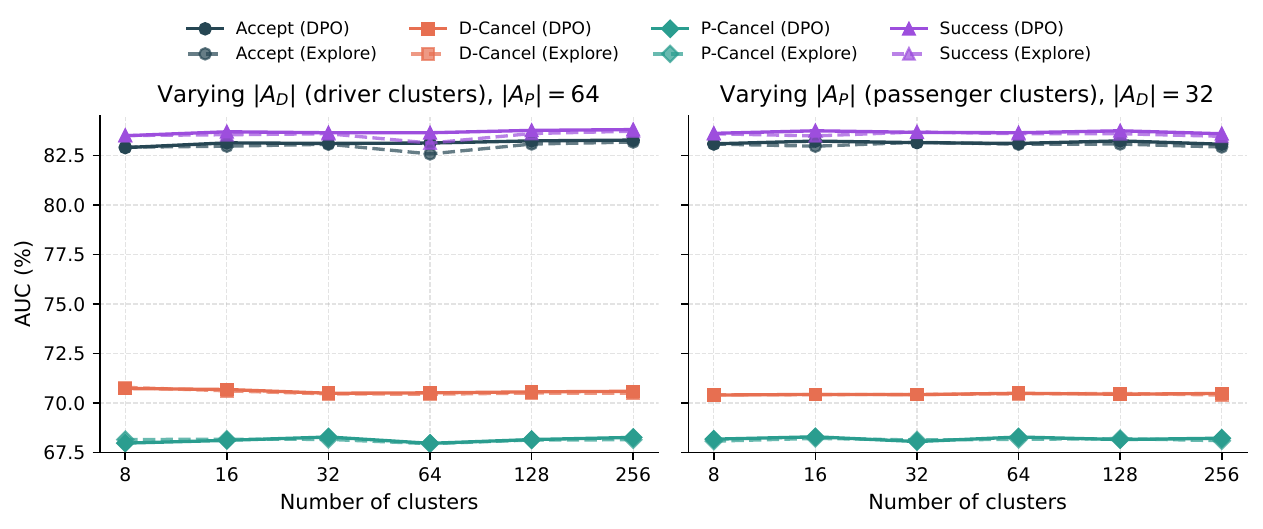}
\vspace{-0.5cm}
\caption{Cluster-count sensitivity for ProfiLLM (dashed) and ProfiLLM-DPO (solid) on City~A, with each curve aggregated over 9 LLM backbones. Left: varying $|A_D|$ with $|A_P|=64$. Right: varying $|A_P|$ with $|A_D|=32$. The cross-model standard deviation across the 9 backbones is 0.05\%--0.27\%.}
\Description{Two side-by-side line plots showing prediction AUC for the four outcome tasks as the number of driver or passenger clusters is swept across 8, 16, 32, 64, 128, and 256. ProfiLLM (dashed) and ProfiLLM-DPO (solid) curves rise from 8 to 16 clusters and then plateau, with all four metrics remaining within a 0.6 percentage-point band beyond 16 clusters.}
\label{fig:cluster_sens}
\end{figure}

Figure~\ref{fig:cluster_sens} shows that both ProfiLLM and ProfiLLM-DPO produce stable AUC across the entire sweep. Performance climbs from 8 to 16 clusters per side and then plateaus, with variation within 0.6\,p.p.\ absolute AUC for all four tasks. Across all 11 cluster configurations, both ProfiLLM and ProfiLLM-DPO outperform the structured-only baseline on Accept, D-Cancel, and P-Cancel at every cluster count; among the 99 runs the un-aligned baseline backbones remain mixed (consistent with Table~\ref{tab:auc_results}), confirming that the ProfiLLM gains observed in the main paper are not specific to a particular cluster count. Across the same sweep, the two variants are statistically indistinguishable on prediction AUC: average $\Delta=$ +0.14\% (Accept), +0.05\% (D-Cancel), +0.02\% (P-Cancel), and +0.14\% (Success), with the sign of $\Delta$ fluctuating across configurations rather than systematically favoring either variant. This supports our deployment choice of ProfiLLM-DPO, which achieves comparable prediction quality at substantially lower offline refresh cost (Appendix~\ref{app:system_cost}).

\section{Utility-Proxy Sensitivity to the Blending Coefficient $\lambda$}
\label{app:lambda}

The blending coefficient $\lambda$ in Eq.~\eqref{eq:fused} controls how strongly the LOGIC rules are mixed with the base production model during \emph{offline profile evaluation}. It does not appear in the online prediction model. We perform a grid search over $\lambda\in\{0,0.1,\ldots,1.0\}$ across 6 cluster granularities for each of 3 outcome tasks, yielding 494 cluster-task combinations (21 driver-accept, 82 driver-cancel, 391 passenger-cancel clusters; counts vary because clusters with insufficient behavioral data are filtered out).

\begin{figure}[t]
\centering
\includegraphics[width=\linewidth]{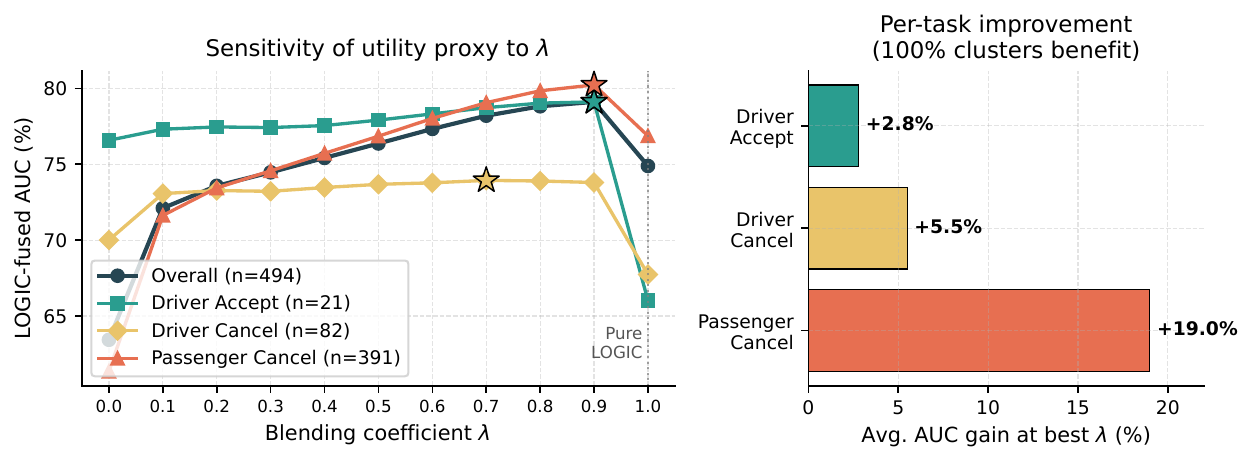}
\vspace{-0.5cm}
\caption{Sensitivity of the LOGIC-rule utility proxy to the blending coefficient $\lambda$. Left: per-task LOGIC-fused AUC versus $\lambda\in[0,1]$, averaged across 494 cluster-task combinations; stars mark each task's peak. Right: average AUC gain at the best $\lambda$ for each task. $\lambda=1.0$ denotes pure LOGIC with the base model discarded.}
\Description{Two side-by-side plots characterising sensitivity to the LOGIC blending coefficient lambda. The left line plot shows AUC versus lambda in [0,1] for Driver Accept, Driver Cancel, and Passenger Cancel, with stars at each task's peak; all tasks rise from lambda equals 0, peak between 0.7 and 0.9, then drop sharply at 1.0. The right bar chart shows the average AUC gain at each task's best lambda, with the largest gain for Passenger Cancel.}
\label{fig:lambda}
\vspace{-0.5cm}
\end{figure}

Figure~\ref{fig:lambda}~(left) reports the average LOGIC-fused AUC as a function of $\lambda$. Three patterns are clear:
(1)~All tasks improve over $\lambda=0$ (base model only) once blending is introduced, demonstrating that LOGIC rules supply \emph{complementary} discriminative signal beyond structured features.
(2)~All tasks collapse at $\lambda=1.0$ (pure LOGIC, base model discarded), confirming the conservative design that retains the strong production predictor.
(3)~The optimum is task-dependent: Driver Accept and Passenger Cancel peak at $\lambda=0.9$, whereas Driver Cancel peaks at $\lambda=0.7$--$0.8$ and stays remarkably flat over $\lambda\in[0.1,0.9]$. At the cluster level, 65.7\% of passenger clusters prefer $\lambda=0.9$ while 40.2\% of driver-cancel clusters prefer $\lambda=0.1$, motivating adaptive per-cluster $\lambda$ selection as a direction for future work.

Figure~\ref{fig:lambda}~(right) summarises the per-task gain at each task's best $\lambda$: +2.8\% (Driver Accept), +5.5\% (Driver Cancel), and +19.0\% (Passenger Cancel). The cancellation tasks gain the most, mirroring the prediction-AUC pattern in Table~\ref{tab:auc_results}: behavioral profiling shines exactly where structured features struggle most, namely in capturing the contextual decision logic behind cancellations.

\section{DPO vs.\ Exploration: Why Both Variants Help}
\label{app:dpo_vs_explore}

A natural question is why we report both ProfiLLM (exploration only) and ProfiLLM-DPO (with DPO fine-tuning) when their per-task prediction AUCs in Table~\ref{tab:auc_results} are similar. The two variants are complementary by design, and we clarify their roles below.

\textbf{(1) DPO targets generator efficiency, not per-task AUC.}
ProfiLLM (exploration) generates $K=5$ candidate profiles per cluster and refines the best for $T=3$ iterations. The theoretical upper bound is $K(1+T)=20$ LLM calls per cluster (5 initial + 5 re-explored per refinement iteration), which we cap at 15 in deployment for compute efficiency. ProfiLLM-DPO generates a high-quality profile in a \emph{single pass}, eliminating the iterative search. When profiles are refreshed for new clusters, new cities, or updated behavioral data, this collapses the offline LLM-call budget by an order of magnitude (Appendix~\ref{app:system_cost}). Both variants serve identically online via cached embeddings; the difference is purely offline.

\textbf{(2) The two variants achieve comparable prediction quality across configurations.}
Across the 11 cluster configurations of Appendix~\ref{app:cluster_sensitivity}, the average inter-variant gap is only +0.14\% (Accept AUC), +0.02\% (P-Cancel AUC), and +0.14\% (Success AUC), with the sign of the gap fluctuating across configurations rather than systematically favoring either variant (maximum absolute deviation is 0.55\,p.p.). Both variants substantially outperform all 7 baseline LLMs in every metric and every city (Tables~\ref{tab:results}--\ref{tab:auc_results}), so the choice between them is dominated by cost, not quality.

\textbf{(3) Per-task differences in Table~\ref{tab:auc_results} reflect different optimization strategies, not degradation.}
ProfiLLM (exploration) performs \emph{per-cluster local optimization}: it iteratively searches profile space for each cluster, and the LOGIC-rule AUC proxy directly drives the selection. ProfiLLM-DPO aggregates preference pairs \emph{across clusters} and learns a single generator (Qwen3-8B) that produces high-utility profiles in one pass. The latter trades a small amount of per-cluster local optimality for cross-cluster generalization. A second, subtler factor is the \emph{signal-channel gap}: DPO is supervised on LOGIC-rule AUC (a discrete Boolean projection), while the downstream prediction model consumes PROFILE \emph{text embeddings} (continuous dense vectors). The two share the same behavioral understanding but are not identical, so DPO optimization on one does not transfer perfectly to the other. Finally, online GMV is a composite matching objective over all three predicted outcomes, so small per-task differences can cancel or compound through the matching weights.

We therefore deploy ProfiLLM-DPO in production because it achieves comparable prediction quality at substantially lower offline refresh cost, enabling faster iteration when scaling to new clusters and cities.

\section{Offline System Cost Analysis}
\label{app:system_cost}

Table~\ref{tab:cost} reports the end-to-end offline cost of running ProfiLLM on a single city with $|A_D|=32$ driver clusters and $|A_P|=64$ passenger clusters using Gemini-3-Pro as the analyst LLM (input \$1.25/1M tokens, output \$10.00/1M tokens). Downstream model training uses one NVIDIA L20 GPU (48GB) at \$1.50/GPU-hour.

\begin{table}[t]
\centering
\caption{Offline pipeline cost breakdown for one city. The Profile + Exploration LLM-call count is $(32+64)\times 15=1{,}440$, where 15 is the deployed per-cluster cap (theoretical max $K(1+T)=20$ with $K=5, T=3$).}
\label{tab:cost}
\footnotesize
\setlength{\tabcolsep}{3.2pt}
\renewcommand{\arraystretch}{1.05}
\resizebox{\linewidth}{!}{%
\begin{tabular}{@{}l c c c c r@{}}
\toprule
\textbf{Stage} & \textbf{Hardware} & \textbf{Wall Time} & \textbf{LLM Calls} & \textbf{Tokens (in/out)} & \textbf{Cost} \\
\midrule
\multicolumn{6}{l}{\textit{Initial run (ProfiLLM, exploration)}} \\
Global Knowledge Mining   & CPU+API      & $\sim$50\,min  & $\sim$20    & 2M/0.5M  & \$7.50 \\
Profile + Exploration     & CPU+API      & $\sim$240\,min & $\sim$1{,}440 & 12M/3M   & \$45.00 \\
Downstream Training       & 1$\times$L20 & $\sim$85\,min  & 0           & --        & \$2.13 \\
\textbf{Total initial}    & --           & \textbf{$\sim$6.3\,hrs} & \textbf{$\sim$1{,}460} & 14M/3.5M & \textbf{\$54.63} \\
\midrule
\multicolumn{6}{l}{\textit{Subsequent refresh (ProfiLLM-DPO, single-pass)}} \\
Profile Generation        & CPU+API      & $\sim$25\,min  & 96          & 0.8M/0.2M & \$3.00 \\
Downstream Training       & 1$\times$L20 & $\sim$85\,min  & 0           & --        & \$2.13 \\
\textbf{Total refresh}    & --           & \textbf{$\sim$1.8\,hrs} & \textbf{96} & 0.8M/0.2M & \textbf{\$5.13} \\
\bottomrule
\end{tabular}}
\end{table}

\begin{figure}[t]
\centering
\includegraphics[width=\linewidth]{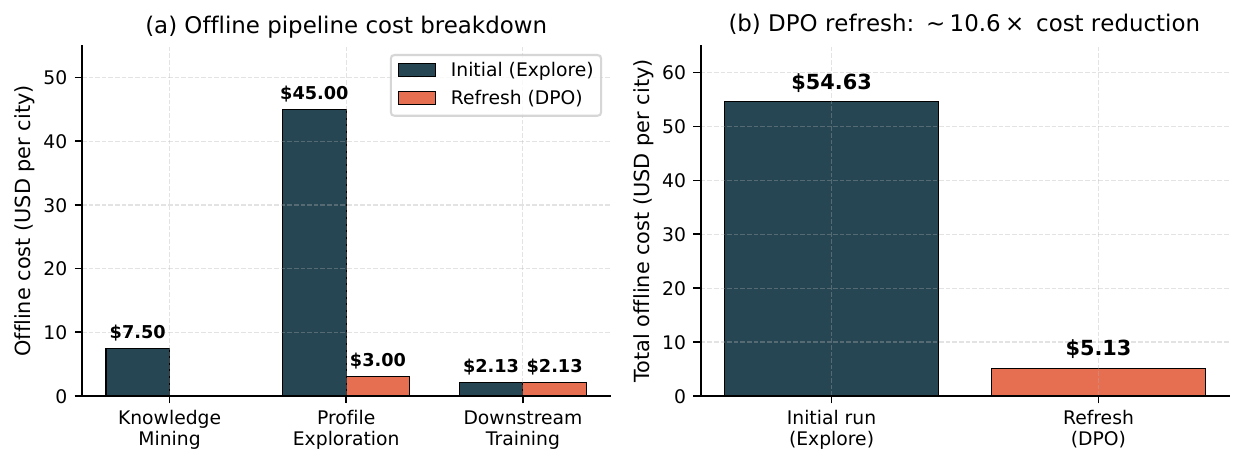}
\vspace{-0.5cm}
\caption{Offline cost breakdown. (a) Per-stage cost for the initial ProfiLLM run vs.\ a single-pass DPO refresh. (b) Total per-city offline cost for the two settings.}
\Description{A two-panel cost breakdown chart. Panel (a) is a stacked bar chart comparing per-stage offline cost (knowledge mining, profile exploration, embedding) between the initial ProfiLLM run and a single-pass DPO refresh, with exploration dominating the initial run and shrinking dramatically under DPO. Panel (b) is a bar chart of total per-city offline cost, showing the DPO refresh reduces total cost by approximately 10.6 times relative to the initial ProfiLLM run.}
\label{fig:cost}
\vspace{-0.5cm}
\end{figure}

Three observations follow (Figure~\ref{fig:cost}).
(1)~\textbf{Cluster-level profiling drives cost efficiency}: 96 cluster-level profiles cover all 348{,}464 users in City~A, a $3{,}630\times$ reduction over per-user profiling and the precondition for affordable LLM-driven profiling at platform scale.
(2)~\textbf{DPO compounds the efficiency gain}: once DPO is trained, subsequent refreshes require only 96 single-pass calls, reducing per-city LLM cost from \$52.50 to \$3.00 and total refresh cost from \$54.63 to \$5.13 ($\sim 10.6\times$ reduction).
(3)~\textbf{Online overhead is negligible}: at serving time the system performs only deterministic cluster assignment ($<$0.01\,ms) and a cached embedding lookup ($<$0.001\,ms), well within DiDi's 200\,ms latency budget; no LLM is queried online.

The 14-day A/B improvement of +0.47\% GMV on a platform processing millions of daily orders translates into revenue gains that exceed the offline cost by several orders of magnitude, even before considering the platform-side savings from reduced cancellations and bad-experience rates.

\section{Complexity Analysis}
\label{app:complexity}

We analyze the computational complexity of ProfiLLM along offline, online, and storage dimensions, and compare against per-user profiling. Notation follows Section~\ref{sec:problem}: $M=|\mathcal{P}|$ passengers and $N=|\mathcal{D}|$ drivers; $\mathcal{H}=\bigcup_u \mathcal{H}_u$ aggregated history with $|\mathcal{H}|=\sum_u |\mathcal{H}_u|$; $\mathcal{A}=\mathcal{A}_D\cup\mathcal{A}_P$ cluster set; $K{=}5$ candidates per generation, $T{=}3$ refinement iterations; $d{=}768$ embedding dimension; and $|\mathcal{C}|$ candidate OD pairs per dispatching cycle.

\subsection{Offline Complexity}

\textbf{(O1) Knowledge Mining (Algorithm~\ref{alg:mining}).}
Each tool performs at most a single pass over $\mathcal{H}$ at $O(|\mathcal{H}|)$ cost. The Explore phase invokes the basic tool set $\mathcal{T}_{\textit{basic}}\subset\mathcal{T}$; Deepen and Validate may then invoke any tool in $\mathcal{T}$ via targeted chains, with each tool used at most once across the workflow (so the total tool-invocation count is $O(|\mathcal{T}|)$). The aggregate tool-execution cost is therefore $O(|\mathcal{T}|\cdot|\mathcal{H}|)$. The LLM agent issues a constant number of reasoning calls (${\approx}20$ in deployment, Table~\ref{tab:cost}), independent of $|\mathcal{H}|$. This stage is linear in data and constant in LLM calls.

\textbf{(O2) Cluster Assignment.}
For each user $u$, we evaluate $|\mathcal{A}|$ membership rules over $\mathcal{H}_u$, costing $O(|\mathcal{A}|\cdot|\mathcal{H}_u|)$. Summed over all users this is $O(|\mathcal{A}|\cdot|\mathcal{H}|)$ and is embarrassingly parallel.

\textbf{(O3) Profile Exploration (Algorithm~\ref{alg:profile}).}
Per cluster $a$ with aggregated history $\mathcal{H}_a$ (where $\sum_a |\mathcal{H}_a|\le|\mathcal{H}|$), the initial generation issues $K$ LLM calls, and each of the $T$ refinement iterations regenerates $K$ candidates conditioned on prediction-error feedback, contributing $KT$ additional calls; every generated candidate's LOGIC rule is then evaluated over $\mathcal{H}_a$ at $O(|\mathcal{H}_a|)$ cost. Aggregating over clusters:
\[
\text{LLM calls}=O\bigl(|\mathcal{A}|\cdot K(1+T)\bigr),\qquad \text{LOGIC eval}=O\bigl(K(1+T)\cdot|\mathcal{H}|\bigr).
\]
The theoretical upper bound is $K(1+T)=20$ calls per cluster; in deployment we cap total calls at $15$ per cluster, which corresponds to early-terminating refinement after at most two iterations of the $K$-candidate regeneration, yielding $|\mathcal{A}|\cdot 15=1{,}440$ calls per city for $|\mathcal{A}|=96$ (Table~\ref{tab:cost}). This is the LLM-call-dominated stage, but the cost is amortized across all users in each cluster.

\textbf{(O4) DPO Fine-tuning.}
Preference-pair construction over the $K(1+T)$ profiles per cluster is at most $O\bigl(|\mathcal{A}|\cdot K^2(1+T)^2\bigr)$. The DPO training cost is the standard LLM fine-tuning loop, $O(E\cdot|\mathcal{P}_{\textit{pref}}|\cdot L\cdot c_{\textit{LLM}})$, where $E$ is the number of epochs, $L$ is profile token length, and $c_{\textit{LLM}}$ denotes the per-token forward+backward FLOPs of the base model. In our deployment this takes ${\approx}85$\,minutes on one NVIDIA L20 GPU.

\textbf{(O5) Embedding Precomputation.}
A single encoder forward per cluster profile: $O(|\mathcal{A}|\cdot L\cdot c_{\textit{enc}})$. Empirically negligible (seconds for $|\mathcal{A}|{=}96$).

\subsection{Online Complexity}

\textbf{(N1) Per OD-pair scoring.}
Passenger and driver cluster IDs are pre-assigned offline and refreshed at user registration; serving requires only two $O(1)$ cache lookups returning $\mathbf{e}_p,\mathbf{e}_d\in\mathbb{R}^d$, an $O(d)$ feature concatenation, and a prediction-network forward $O(c_f)$.

\textbf{(N2) Per-cycle cost.}
For $|\mathcal{C}|$ candidate OD pairs the per-cycle cost is $O\bigl(|\mathcal{C}|\cdot(d+c_f)\bigr) \;+\; \mathcal{O}_{\text{KM}}(|\mathcal{C}|)$,
where $\mathcal{O}_{\text{KM}}$ denotes the production Lagrangian-relaxation Kuhn--Munkres matching cost (identical to the structured-only baseline; see Appendix~\ref{app:simulator}). Profile features add only the $O(|\mathcal{C}|\cdot d)$ feature-concat term, which empirically contributes well under $1$\,ms per cycle (Appendix~\ref{app:system_cost}).

\textbf{(N3) Cold-start.}
Users without sufficient history are mapped to a default cluster at registration ($O(1)$); no additional online cost.

\subsection{Storage Complexity}

ProfiLLM's serving state comprises three components. The cluster embeddings occupy $|\mathcal{A}|\cdot d\cdot 4$\,B $=96\times768\times 4\approx 295$\,KB; the user-to-cluster table stores a 32-bit cluster ID per user at 4\,B each, $\approx 1.4$\,MB for the $348{,}464$ users in City~A; and the LOGIC rules and PROFILE text amount to $|\mathcal{A}|$ short strings, $\approx 50$\,KB.
The active footprint is therefore a few MB per city, nearly three orders of magnitude smaller than caching a per-user $d$-dimensional embedding.

\subsection{Comparison with Per-User Profiling}

Cluster-level profiling reduces both LLM-call count and embedding storage by a factor of $(M+N)/|\mathcal{A}|$. For City~A:
\[
\frac{|\mathcal{P}\cup\mathcal{D}|}{|\mathcal{A}|}=\frac{348{,}464}{96}\approx 3{,}630\times.
\]
\noindent(The deployment registry contains $348{,}464$ users; Appendix~\ref{app:heterogeneity} reports $345{,}294$ users active within a narrower 38-day analysis window, hence the small discrepancy.)
This is the structural source of ProfiLLM's offline cost efficiency, and Appendix~\ref{app:cluster_sensitivity} confirms that this reduction does not sacrifice prediction quality once $|\mathcal{A}|>16$.

\subsection{Summary}

Table~\ref{tab:complexity} consolidates the analysis. Offline complexity is linear in data ($|\mathcal{H}|$) for tool execution and cluster assignment, and the LLM-call count is linear in the \emph{cluster} count $|\mathcal{A}|$ rather than the \emph{user} count $(M+N)$. Online complexity is dominated by the existing bipartite-matching solver; profile features add only $O(|\mathcal{C}|\cdot d)$ FLOPs and two $O(1)$ cache lookups per OD pair. Storage for LLM-introduced artifacts is sub-MB.

\begin{table}[t]
\centering
\footnotesize
\setlength{\tabcolsep}{4pt}
\renewcommand{\arraystretch}{1.1}
\caption{Complexity summary. $|\mathcal{H}|$: total history records; $|\mathcal{A}|$: cluster count; $K{=}5,T{=}3$; $|\mathcal{C}|$: candidate OD pairs per cycle; $d{=}768$; $c_f$: constant prediction-network forward cost.}

\begin{tabular}{@{}lll@{}}
\toprule
\textbf{Stage} & \textbf{Time} & \textbf{Notes} \\
\midrule
Knowledge Mining (O1)    & $O(|\mathcal{T}|\cdot|\mathcal{H}|)$               & $+\,O(1)$ LLM calls \\
Cluster Assignment (O2)  & $O(|\mathcal{A}|\cdot|\mathcal{H}|)$               & parallel \\
Profile Exploration (O3) & $O(K(1+T)\cdot|\mathcal{H}|)$                      & $+\,O(|\mathcal{A}|\cdot K(1+T))$ LLM calls \\
DPO Training (O4)        & $O(E\cdot|\mathcal{P}_{\textit{pref}}|\cdot L)$    & one-time \\
Embedding (O5)           & $O(|\mathcal{A}|\cdot L)$                          & seconds \\
\midrule
Online per OD pair (N1)  & $O(d+c_f)$                                         & two cache lookups \\
Online per cycle (N2)    & $O(|\mathcal{C}|\cdot d)+\mathcal{O}_{\text{KM}}$   & KM dominates \\
\midrule
Embeddings storage       & $O(|\mathcal{A}|\cdot d)$                          & $\approx 295$\,KB \\
User-cluster table       & $O(M+N)$                                           & $\approx 1.4$\,MB (City~A) \\
\bottomrule
\end{tabular}
\label{tab:complexity}
\end{table}

\section{Extended 14-day A/B Test: Long-term Stability}
\label{app:extended_ab}

The 14-day deployment in Section~\ref{subsec:deployment} extends the initial 5-day pilot to a longer window for more stable estimates. We summarize the comparison and broader generalization evidence here.

\textbf{Stability over time.}
GMV improvement \emph{grew} from +0.36\% (5 days) to \textbf{+0.47\%} (14 days) and CR rose from +0.16\% to +0.33\%, while every cancellation and bad-experience metric remained directionally negative across the full window (CBA, PCR, DCR, BER). The fact that effect sizes did not decay, and in several cases grew, over the extended observation period supports the interpretation that ProfiLLM produces durable matching-quality gains rather than transient effects.

\textbf{Joint movement across funnel stages.}
Every monitored realized rate moves in the beneficial direction across mechanistically distinct stages of the fulfillment funnel: revenue/completion (GMV, CR), pre-acceptance attrition (CBA), post-acceptance cancellation (PCR, DCR), and completed-order experience (BER). Each stage reflects a different behavioral mechanism, so the joint consistency provides converging evidence that the improvement is systematic. A model that improves only one stage would be expected to show mixed signs elsewhere; the uniform direction here is consistent with profiling improving the underlying outcome prediction that feeds every stage.

\textbf{Generalization beyond a single city.}
While the A/B was conducted in City~A, the offline simulator evaluation in Table~\ref{tab:results} covers three cities with distinct supply-demand regimes (City~A supply-constrained, City~B supply-relaxed, City~C large-scale high-demand) and uses the production routing API for realistic ETA. ProfiLLM dominates baselines across all three cities and all time-of-day buckets (Figure~\ref{fig:hourly_performance}), supporting that the online gains would carry over. Broader online rollout across additional cities is in progress and will be reported in a follow-up.

\section{Privacy and Fairness Considerations}
\label{app:ethics}

Because user profiles directly influence which drivers receive which orders, we discuss privacy and fairness implications of ProfiLLM.

\textbf{Privacy by architectural design.}
Two safeguards limit exposure of individual data.
(1)~\textit{Cluster-level abstraction.} The LLM never sees a single user's identifiable trajectory in isolation. It processes only cluster-pooled, re-sampled order records together with summary statistics aggregated over all members of a cluster (\eg, this cluster of drivers cancels orders with pickup distance $>5$\,km during evening hours at rate $r$). An individual user's data only contributes to a cluster's aggregate and cannot be reconstructed from the profile description.
(2)~\textit{Offline-only LLM inference.} Every LLM call happens in the offline knowledge-mining and profile-exploration stages. At serving time the online system performs only a deterministic cluster-assignment rule evaluation and a cache lookup for pre-computed embeddings (Section~\ref{subsec:prediction}); no user data is transmitted to any LLM API during real-time dispatching.

\textbf{Driver earning equity.}
If certain driver clusters are profiled as ``likely to cancel long-pickup orders,'' the system may avoid such assignments, improving platform efficiency but potentially affecting those drivers' earning opportunities. This is, however, the behavior any accurate prediction model would produce, whether using profiles or structured features; ProfiLLM merely makes the prediction signal more explicit and \emph{auditable}. Platform operators can inspect the textual PROFILE of each cluster (Section~\ref{subsubsec:candidate}) and verify it does not encode undesirable biases, a transparency advantage over opaque deep-feature interactions.

\textbf{Passenger service equity.}
Infrequent users, including the 96\% long-tail passengers in Figure~\ref{fig:longtail}, are still assigned to behavioral clusters via $a^*(u)$ and receive shared cluster-level profiles rather than being excluded from profiling (Section~\ref{subsubsec:integration}); only genuine cold-start users with no usable history fall back to a default cluster. They therefore receive at least baseline matching quality, closing the cold-start gap that per-user profiling methods would face.

\textbf{Behavioral vs.\ protected attributes.}
The clustering rules use only behavioral features (order patterns, cancellation history, temporal activity, spatial preferences), never protected attributes. We acknowledge that behavioral features may correlate with such attributes (commute-hour patterns with occupation, frequent regions with socioeconomic status). The cluster-level design supports a tractable auditing path: outcome distributions (driver income, passenger wait time, allocation rates) can be measured across clusters and clusters with statistically anomalous treatment flagged. We are integrating such auditing into our deployment pipeline as an ongoing direction.

\section{Prompt Template}
We present the abstracted prompt template in Table~\ref{tab:prompt}.
\balance  

\begin{table*}[b]
\centering
\caption{Prompt templates used for profile exploration. Placeholders denote injected inputs; the concrete feature legend and data tables are omitted.}
\label{tab:profile-optimization-prompts}
\setlength{\tabcolsep}{7pt}
\renewcommand{\arraystretch}{1.15}
\rowcolors{3}{black!3}{white}
\begin{tabularx}{\textwidth}{@{}lX@{}}
\toprule
\textbf{Template} & \textbf{Abstracted Prompt} \\
\midrule

Driver---Draft &
\promptcell{
\pcode{[Role]} Expert analyst (data science + behavioral economics) for ride-hailing driver behavior. \par
\pcode{[Task]} Infer a stable driver persona and decision logic. \par
\pcode{[Inputs]} Summary: \ph{SUMMARY\_STATS}; Grouped recent records: \ph{RECENT\_GROUPED\_RECORDS}. \par
\pcode{[Guidelines]} Records may be re-sampled; focus on feature differences across outcomes; use only features in the provided legend. \par
\pcode{[Reasoning]} (1) rejection patterns (ignored) (2) regret patterns (post-accept cancellations) (3) persona + generalizable rules. \par
\pcode{[Output]} XML only: \pcode{<ANALYSIS>}, \pcode{<PROFILE>}, \pcode{<LOGIC\_ACCEPT>} (1-line Python), \pcode{<LOGIC\_CANCEL>} (1-line Python).
} \\

Driver---Improve &
\promptcell{
\pcode{[Inputs]} \ph{SUMMARY\_STATS}, \ph{RECENT\_GROUPED\_RECORDS}, plus previous response \ph{PREVIOUS\_RESPONSE} and feedback \ph{FEEDBACK}. \par
\pcode{[Task]} Improve profile + logic to increase validation performance; change only what is justified by patterns and feedback. \par
\pcode{[Output]} A complete, self-contained updated response (not a patch), in the same XML-only format.
} \\

Passenger---Draft &
\promptcell{
\pcode{[Role]} Expert analyst for ride-hailing passenger post-match behavior. \par
\pcode{[Task]} Infer patience and post-match cancellation triggers. \par
\pcode{[Inputs]} Summary: \ph{SUMMARY\_STATS}; Grouped recent records (completed vs cancelled-after-match): \ph{RECENT\_GROUPED\_RECORDS}. \par
\pcode{[Guidelines]} Compare feature differences across outcomes; use only features in the provided legend (high-level: price/trip, ETA/waiting, context such as time and weather). \par
\pcode{[Reasoning]} (1) time vs money (2) sunk cost (3) context modifiers (4) persona synthesis. \par
\pcode{[Output]} XML only: \pcode{<ANALYSIS>}, \pcode{<PROFILE>}, \pcode{<LOGIC\_CANCEL>} (1-line Python).
} \\

Passenger---Improve &
\promptcell{
\pcode{[Inputs]} \ph{SUMMARY\_STATS}, \ph{RECENT\_GROUPED\_RECORDS}, plus previous response \ph{PREVIOUS\_RESPONSE} and feedback \ph{FEEDBACK}. \par
\pcode{[Task]} Improve profile + logic with minimal, data-justified changes. \par
\pcode{[Output]} A complete, self-contained updated response (not a patch), in the same XML-only format.
} \\

\bottomrule
\label{tab:prompt}
\end{tabularx}
\end{table*}

\end{document}